\documentclass[sigconf, preprint]{acmart}

\AtBeginDocument{%
  \providecommand\BibTeX{{%
    \normalfont B\kern-0.5em{\scshape i\kern-0.25em b}\kern-0.8em\TeX}}}

\setcopyright{acmcopyright}
\copyrightyear{2025}
\acmYear{2025}
\acmDOI{XXXXXXX.XXXXXXX}

\acmConference[KDD '25]{}{August 3--7,
2025}{Toronto, Canada.}
\acmPrice{15.00}
\acmISBN{978-1-4503-XXXX-X/18/06}

\usepackage{algorithm}
\usepackage{algorithmic}
\usepackage{booktabs}
\usepackage{subfigure}
\usepackage{longtable}

\begin{document}

\title{Feature Selection for Network Intrusion Detection}


\author{Charles Westphal}
\email{charles.westphal.21@ucl.ac.uk}
\affiliation{%
  \institution{University College London}
  \department{Department of Computer Science}
  \streetaddress{Gower Street}
  \city{London}
  \country{UK}
  \postcode{WC1E 6BT}
}

\author{Stephen Hailes}
\email{s.hailes@ucl.ac.uk}
\affiliation{%
  \institution{University College London}
  \department{Department of Computer Science}
  \streetaddress{Gower Street}
  \city{London}
  \country{UK}
  \postcode{WC1E 6BT}
}

\author{Mirco Musolesi}
\email{m.musolesi@unibo.it}
\affiliation{%
  \institution{University College London}
  \department{Department of Computer Science}
  \streetaddress{Gower Street}
  \city{London}
  \country{UK}
  \postcode{WC1E 6BT}
}

\affiliation{%
  \institution{University of Bologna}
  \department{Department of Computer Science}
  \streetaddress{Via Forzoni, 35}
  \city{Bologna}
  \country{Italy}
  \postcode{40126}
}

\newcommand{\cw}[1]{\footnote{\textbf{Charlie: #1}}}
\newcommand{\sh}[1]{\footnote{\textbf{Steve: #1}}}
\newcommand{\mm}[1]{\footnote{\textbf{Mirco: #1}}}

\renewcommand{\shortauthors}{Trovato and Tobin, et al.}

\begin{abstract}
  Network Intrusion Detection (NID) remains a key area of research within the information security community, while also being relevant to Machine Learning (ML) practitioners. The latter generally aim to detect attacks using network features, which have been extracted from raw network data typically using dimensionality reduction methods, such as principal component analysis (PCA). However, PCA is not able to assess the relevance of features for the task at hand. Consequently, the features available are of varying quality, with some being entirely non-informative. From this, two major drawbacks arise. Firstly, trained and deployed models have to process large amounts of unnecessary data, therefore draining potentially costly resources. Secondly, the noise caused by the presence of irrelevant features can, in some cases, impede a model's ability to detect an attack. In order to deal with these challenges, we present Feature Selection for Network Intrusion Detection (FSNID) a novel information-theoretic method that facilitates the exclusion of non-informative features when detecting network intrusions. The proposed method is based on function approximation using a neural network, which enables a version of our approach that incorporates a recurrent layer. Consequently, this version uniquely enables the integration of temporal dependencies. Through an extensive set of experiments, we demonstrate that the proposed method selects a significantly reduced feature set, while maintaining NID performance. Code will be made available upon publication.
\end{abstract}




\received{20 February 2007}
\received[revised]{12 March 2009}
\received[accepted]{5 June 2009}

\maketitle

\section{Introduction}\label{sec:introduction}

Network Intrusion Detection (NID) remains a key focus of the information security community given its substantial economic impact. For example, IBM estimates showed that, in 2023, the average cost of a data breach for the afflicted party was USD 4.45 million \cite{ibmreport2023}.
During standard network operations malicious attacks are typically absent. Therefore, the earliest Intrusion Prevention Systems (IPSs) were statistical methods developed to detect irregularities in network data \cite{denning87}. These methods were among the early implementations of anomaly detectors, a broader category of techniques that continues to be effective in NID \cite{lazarevic2003,jabez15}. Despite their success \cite{ahmed2016}, traditional anomaly detection methods, primarily focused on single time series analysis, fall short in leveraging the interrelations among multiple data series. In contrast, anomaly detection methods designed to exploit these inter-series relationships often face significant computational complexity challenges \cite{dobigeon07,harle14}.

Starting from these considerations, NID has also been studied as a tabular data classification problem \cite{kevric2017,javaid2016,subba2016}. According to this paradigm, a machine learning (ML) model receives as input a vector that summarizes the network data at a given instance in time, and outputs a boolean or probability indicating whether or not the system is under attack, and, in some cases, what type of attack it is under. Usually, the vector of features is extracted from raw network data, which are collected and stored as packet capture (PCap) files. These can be represented as a highly-dimensional set of time-series. A dimensionality-reduction algorithm is then applied to extract individual features. Examples of techniques employed for this task include principal component analysis (PCA) \cite{pearson1901,purnama20}, Linear Discriminant Analysis (LDA) \cite{lachenbruch1979,chen23} and, more recently, deep learning-based approaches \cite{sun2020}. Although these methods have the ability to cluster the data into features, they are unable to assess how relevant and informative they will be for classification. Consequently, some will be entirely non-informative and should be removed. Additionally, given the NID context, it is possible to outline three further reasons as to why one may wish to reduce the required feature data. Firstly, the acquisition of the network data may be costly since specific software (and, in some cases, hardware) has to be deployed. Secondly, measurements might introduce delays into the network operations. Finally, in the context of cybersecurity, minimizing measurements is of strategic importance. Since in certain situations, an advantage can be gained by giving the impression that the network is not being monitored at all.

This has led to a growing body of work in feature selection for NID \cite{khammassi17,moustafa17,selvakumar19,khammassi20}. 
It is possible to identify two groups of approaches. The first aims to maximize mutual information (MI), i.e., the correlation, between the chosen feature set and the attack vector labels \cite{peng05,gao16,brown12,chen18}. However, this requires the definition of the desired set size. In feature-rich datasets, including those reporting network traffic, the search space of this hyperparameter becomes impractically large. Moreover, the complexity of these algorithms scales polynomially in time with the number of features to be assessed. Again, this is non-desirable for highly dimensional data. The second, and most common, general paradigm is to rank features based on MI \cite{LIU2009}, or Shapley values \cite{Shapley1953}, before selecting a predetermined number of the best performing features. However, these methods struggle to deal with highly correlated features \cite{Frye2020,Kumar2020}. This issue is especially pertinent in network contexts, where homogeneous traffic patterns are repeated in individual features. Additionally, none of the methods discussed so far account for temporal dependencies, which are of key importance in NID. While the use of ML models with recurrent layers has been shown to enhance NID performance, the feature selection methods used to determine inputs for these models fail to leverage the same temporal relationships. Consequently, the set of selected features may lack crucial information that could otherwise be utilized effectively by the classification system.
We identify four major challenges when selecting features for NID:
\begin{itemize}
    \item  \textbf{The complexity problem.} How does the complexity of the method scale with the number of selected features? (In NID where the number of features is large, complex algorithms quickly become impractical.)
    \item  \textbf{The length $k$ problem.} How does the feature selection method define the final number of selected features? (This is of particular relevance for datasets that have large numbers of features as the size of the resulting search space scales proportionally.)
    \item \textbf{The redundancy problem.} How does the feature selection method handle highly correlated variables? (This is key during NID as multiple features portrey similar traffic patterns.)
    \item \textbf{The temporal relationships problem.} Can the feature selection method incorporate temporal dependencies? (In NID incorporating temporal dependencies into classification tasks is known to improve accuracy, therefore, this should also inform the features selected.)
\end{itemize}

In response to these challenges, we present  Feature Selection for Network Intrusion Detection (FSNID), a novel information-theoretic method  that naturally overcomes such problems. FSNID relies upon the ability to neurally approximate the \textit{entropy transferred} from the labels dataset to each feature. Once this quantity is obtained, we then exclude variables from our set of desired features if they are deemed to be uninformative.  Furthermore, given it relies upon function approximation, we can use recurrent layers to incorporate time-based dependencies within our Transfer Entropy (TE) calculation. For example, using a simple DNN to estimate the entropy transferred will lead to a value that only reflects the relationships present in the data at each iteration of training. Meanwhile, if we were to use an Long-Short Term Memory (LSTM) network \cite{hochreiter1997} to approximate this function we can also include relationships that span multiple iterations, overcoming the \textit{temporal relationships problem} presented above. We show that incorporating such dependencies reduces the number of required features further. Consequently, our method reduces the feature space significantly, while not degrading classification performance. The contributions of the paper can be summarized as follows:
\begin{itemize}
\item We develop a method of feature selection for NID, based on excluding variables that \textit{transfer} negligible entropy to the attack vector labels. 
\item The computation of the TE-based measure in FSNID depends on general function approximation methods. By incorporating recurrent layers into this estimation process, we leverage \textit{temporal} dependencies.
\item We show experimentally that using the described techniques leads to significant reductions in the size of the feature space without degrading classification performance.
\end{itemize}

\section{Related Work}\label{sec:rel_work}
Feature selection can be defined as the process of reducing the dimensionality of the input to the ML model while maintaining the same classification or regression performance. Methods for feature selection can be divided into two conceptual groups, namely wrapper and filter methods, where the latter have been deployed more commonly in the field of feature selection for NID due to their performance for high-dimensional feature spaces. 

\noindent \textbf{Shapley value based filter methods.}  These involve selecting variables due to the underlying relationships within the data, independently from the model itself. A popular method relies on Shapley values as a score for `feature importance' \cite{covert2020}. Shapley values, originated by Shapley as a means to efficiently allocate resources in cooperative game theory \cite{Shapley1953}, were adapted for feature importance by having the input features `cooperate' to predict the output of an ML model \cite{Lunberg2017}. 
Despite Shapley value's widespread adoption, they fail to overcome the \textit{redundancy problem}. Two perfectly correlated features, will both return Shapley values of zero, in spite of their importance for predicting the output of the model. Furthermore, \cite{janzing20} and \cite{Sundararajan20} showed that, in some cases, non-negative Shapley values could be assigned to features that have no impact on the final outcome of an ML model. In response to this, literature has been developed which overcomes such issues. Namely, the authors of \cite{janssen2023} and \cite{catav21} developed methods to assign feature importance's that adhere to certain properties, which have been pre-specified in their axioms for feature importance. One such axiom ensures that when calculating the second feature importance of a pair of perfectly correlated features, it must receive the same score as the first. This leads to scores that reflect the importance of a feature in predicting the model output, therefore, negating the redundancy issue as described by \cite{Kumar2020,Frye2020}. However, the use of these scores lead to the following problem: when selecting features for a model, either of the correlated features is just as likely to be selected. Consequently, both are selected. In practice, we require only one of these features. In computer networks, and networks more generally, we often see highly correlated features. Consequently, Shapley-based methods for feature selection are not well suited to network traffic data. 

\noindent \textbf{Information-theoretic filter methods.} As a result, information-theoretic filter methods are more commonly used for NID. These methods aim to exploit underlying relationships in the data to select the features without knowledge of the model in use. A standard method \cite{battiti94} (see also the variations presented in \cite{peng05,brown12}) involves augmenting the feature that maximizes the MI between the chosen subset and the ground truth. However, such methods fail to overcome the \textit{length $k$ problem} and the \textit{complexity problem}. Another related method involves exploring all the permutations of the entire feature set to optimize the MI between the subset and the target variable \cite{gao16,chen18,yamada20}. To manage the computationally intensive task of searching through permutations, such techniques limit the size of the search space by fixing the cardinality of these subsets as \( k \). However, the selected features cannot guarantee optimality, unless all possible values of $k$ are compared. Therefore, these methods also suffer from \textit{the length $k$ problem}. Despite these drawbacks, both techniques represent the state of the art and they are widely adopted by the NID community \cite{amiri11,bostani17,selvakumar19}.

 \noindent \textbf{Feature selection methods for network intrusion detection.} Many specialized feature selection methods for NID have been developed \cite{amiri11,bostani17,khammassi17,moustafa17,selvakumar19,khammassi20}. Often, they ensure that the complexity of the algorithm scales non-polynomially in time with respect to the number of features, but this comes at the cost of robustness. For example, in \cite{chou2008} `irrelevant' features are removed on a case-by-case basis if they are uncorrelated with the target. This is done without considering if these features provide \textit{synergistic} information when considered in groups,  whereas in \cite{selvakumar19}, instead of searching the full space of potential feature subsets, the authors only consider the subsets of size $k$. Therefore, such methods suffer from the length $k$ problem.


\section{Preliminaries}
\subsection{Notation and Terminology}
\label{sec:preliminaries}

We denote sets of random variables using calligraphic symbols (e.g., $\mathcal{X}$), single random variables using capital letters (e.g., $X$), and their realizations with lowercase letters (e.g., $x$). To label a selected feature set we use $\mathcal{X}_*$. The function \( A(\cdot) \) operates on random variables, yielding the set of all their possible realizations. Meanwhile, \( \mathscr{P}(\cdot) \) produces the powerset of its argument. We use Shannon's entropy (e.g., $H(X)$) to represent uncertainties. Let the following list denote the features of a computer network at time $t$, $\xi^t = (x_1^t, x_2^t... x_N^t)$, where $\xi^t$ will act as the input to our classification model. By sampling jointly from the elements of our input list $x_1^t$, and from the realizations of the ground truth $y^t$, we realize not only the set of all possible features as random variables (written as $\mathcal{X} = \{X_1, X_2\dots X_N\}$), but also the ground truth (written as $Y$). 

\subsection{Information-theoretic Concepts}
\label{sec:IT_preliminaries}
We now briefly review the key concepts from Information Theory at the basis of the proposed method.

\noindent \textbf{Mutual information (MI).} The MI is a non-linear extension of the Pearson correlation. Formally it is written as the difference between a conditional and non-condition Shannon's entropy \cite{shannon1948}. For example:
 \begin{equation}
\label{eqn:MI}
    I(X;Y) = H(X) - H(X|Y).
\end{equation}
Equation \ref{eqn:MI} describes the reduction in uncertainty of the observation of $X$ caused by observing $Y$. MI is widely used for feature selection, both when applied to NID \cite{amiri11,bostani17,selvakumar19} and more generally \cite{peng05,brown12,gao16,chen18,covert23}.

\noindent \textbf{Transfer entropy (TE).} TE measures the directed transfer of information between two random processes \cite{schreiber2000}. More formally, we write this quantity as:
 \begin{equation}
\label{eqn:TE}
    TE_{X \rightarrow Y} = H(X^{t+1}|X^{t}) - H(X^{t+1}|X^{t},Y^{t}).
\end{equation}
Equation \ref{eqn:TE} measures the \textit{extra} reduction in uncertainty of the \textit{next} realization of $X$ by considering the current observation of $Y$.  

\noindent \textbf{Redundancy.} Redundancy measures the difference between the maximum and observed uncertainty of an ensemble of random variables. If this difference is large, it implies that many of the variables provide indistinguishable information \cite{Plaus1996}. To explain why this is a key concept in feature selection, we present the following example. Let there exist two variables ($X_i$ and $X_j$) that provide identical information regarding the target ($Y$), more formally $I(X_i;Y)=I(X_j;Y)=I(X_j,X_i;Y)$. An optimal feature selection algorithm will include only one, to reduce the input dimensionality, while maintaining the total information. However, this has been proven difficult to rigorously implement \cite{Frye2020,Kumar2020}. Some methods \cite{breiman2001,debeer20}, assign a feature importance score of zero to both variables, resulting in neither being selected. Conversely, other techniques identify them as equally important and select both \cite{catav21,janssen2023}.

\noindent \textbf{Synergy.} Synergy describes the extra information provided due to variables being considered in combination as opposed to individually. The classic example given of such a relationship is implemented using an XOR function. Let there exist two binary string variables $X_i$ and $X_j$. Furthermore, let us suppose that a third variable $Y$ is calculated as the XOR of these two variables realizations. It follows that $X_i$ and $Y$ and $X_j$ and $Y$ are completely uncorrelated ($I(X_i;Y)=I(X_j;Y)=0$), but when considered together they are fully informative  ($I(X_i, X_j;Y)=H(X)$) \cite{williams2010}. 
Features that are characterized by these relationships are assigned negligible feature importance scores by correlation-based measures (such as those used in \cite{chou2008,brown12,covert23}), despite the information they provide in combination. 

\section{Method}
\subsection{Overview}
We now present our information-theoretic filter method of feature selection for NID. Herein, we consider only filter-based approaches, whether that be FSNID or baselines, due to the advantages they possess for highly dimensional feature spaces (see section \ref{sec:rel_work}).
Initially, we discuss the task of detecting network intrusions using supervised learning. The subsequent part of this section is dedicated to elucidating our feature selection technique.

We start by presenting the mathematical details of the TE based measure that signifies feature importance. Following that, we outline the algorithm we develop to leverage this measure, discussing how it addresses the first three problems highlighted in Section \ref{sec:introduction}. We also present a procedure for estimating the value of the measure. We then describe the method through which we integrate temporal dependencies into the estimation process, thus resolving the issue of temporal relationships.

\subsection{Network Intrusion Detection as Supervised Classification}
In this paper, analogously to \cite{shone2018,jia2019,zeng2019,ahmim2019,verma2019,karatas20}, NID is viewed as a supervised classification task. 
Let us start by considering the case in which all the $N$ features act as inputs to the classifier. They describe the properties of a network at an instance in time, and are characterized by the realizations of $\mathcal{X}$ at time $t$, which we indicate with $\xi^t=(x_1^t, x_2^t, \dots, x_N^t)$. Meanwhile, $y^t$ describes whether or not these properties correspond to benign or malicious traffic. The primary objective of an NID classifier is to infer the value of $y^t$ as output, given $\xi^t$ as input. This process can be conceptualized as a mapping function \( f \), which translates the space of all possible combinations of network properties into the space of all possible attack types, with one such type being the possibility of no attack at all. Formally, this is represented as \( f: A(\mathcal{X}) \mapsto A(Y) \). In this paper, our goal is not only to detect attacks, but also to classify their type.  Consequently, we model $f$ using either a multi-layer-perceptron or a Long-Short Term Memory (LSTM) network with a negative log likelihood loss, where the number of potential outputs is equal to the total number of attack types plus one (the one characterizes benign traffic). 
An optimal function $f^*$ outputs the correct attack label for each iteration (more formally $f^*(\xi^t)=y^t$). The goal of feature selection is to reduce the dimensionality of the input to our classifier, without affecting its performance.

\subsection{Measuring Transfer Entropy}
\label{sec:transfer_entropy_based_measure}
We now define the measure $\Phi_{{X}_i;{\mathcal{X}} \rightarrow {Y}}$, which quantifies the difference in the uncertainty of the target variable, $Y$, when the set $\mathcal{X}$, does and does not include some feature of interest $X_i \in \mathcal{X}$:
 \begin{equation}
\label{eqn:TE_measure}
    \Phi_{{X}_i;{\mathcal{X}} \rightarrow {Y}}  = H({Y}|{\mathcal{X}}_{\backslash {X}_i}) - H({Y}|{\mathcal{X}}).
\end{equation}
This quantity is an adaptation of Schreiber's TE, and describes how features transfer information to the ground truth variable \cite{schreiber2000}. Equation \ref{eqn:TE_measure} measures \textit{the reduction in uncertainty of $Y$'s observations given $X_i$ is added back to $\mathcal{X}$}. Therefore, if $\Phi_{{X}_i;{\mathcal{X}} \rightarrow {Y}} = 0$, adding the variable $X_i$ back to the set $\mathcal{X}$ does not reveal further information about the target variable. The core principle of the methodology is to exclude such variables from the set of features, as they are said to be uninformative regarding the target. This measure has the desirable quality that by considering the effect of \textit{combining} the feature of interest with the remaining variables in $\mathcal{X}$, it natively considers high-order synergistic relationships. However, in the following section, we will illustrate that evaluating this measure for multiple features simultaneously does not overcome the \textit{redundancy problem}, for analogous reasons to the methods developed in \cite{breiman2001}. 

\subsection{Description of the FSNID Algorithm}
\label{sec:application_of_measure}
In this section, we discuss the affects of the \textit{redundancy problem}, before describing the design of the overall FSNID algorithm. We then analyze the performance of FSNID in handling \textit{the complexity} and \textit{length $k$ problems}. 

\begin{figure}[t]
    \subfigure{\includegraphics[width=\columnwidth]{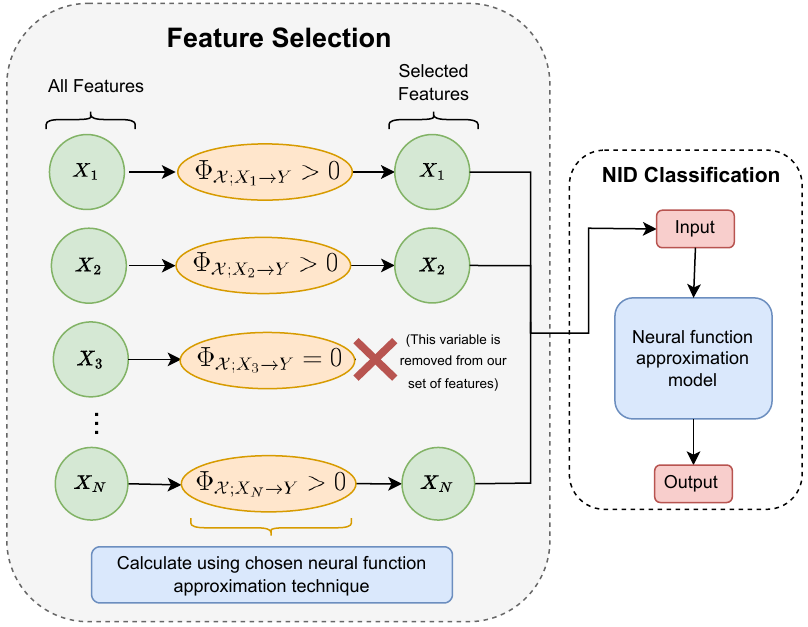} }%
    \caption{Diagrammatic representation of FSNID.}
    \label{fig:schem}
\end{figure}

\noindent \textbf{Dealing with redundancy.} In practice, a simple simultaneous application of Equation \ref{eqn:TE_measure} to all variables fails to overcome the \textit{redundancy problem} \cite{Frye2020,Kumar2020}. To explain this, let us suppose there are two perfectly correlated features ($X_i,X_j$), and we remove either one from the set $\mathcal{X}$ to calculate \(\Phi_{{X}_i;{\mathcal{X}} \rightarrow {Y}}\). All the information that would be lost by removing $X_i$ is replicated in $X_j$ and vice versa. As a result, we calculate \(\Phi_{{X}_i;{\mathcal{X}} \rightarrow {Y}} = 0\) and \(\Phi_{{X}_j;{\mathcal{X}} \rightarrow {Y}} = 0\), and both features are excluded from $\mathcal{X}$, in spite of their ability to reduce the uncertainty of the target. To mitigate the \textit{ redundancy problem}, we apply the measure defined in Equation \ref{eqn:TE_measure} in a sequential manner, as detailed in Algorithm \ref{alg:algorithm} in Appendix \ref{app:fsnid_alg}. Under such circumstances, permanently eliminating variables from the set $\mathcal{X}$ (line 4 of Algorithm \ref{alg:algorithm}), if they satisfy \(\Phi_{{X}_i;{\mathcal{X}} \rightarrow {Y}} = 0\) ensures that, upon encountering the first of the redundant variables, it is removed. As a result, there is no longer redundant information in the set $\mathcal{X}$, and the calculation of the remaining variables' contribution is heavily simplified \cite{gregorutti2017correlation,Frye2020,Kumar2020}. Therefore, this simple method overcomes the \textit{redundancy problem}.
A schematic illustration of this process can be seen in Figure \ref{fig:schem}. 

\noindent \textbf{Dealing with the length $k$ problem and scalability.} It is trivial to see that Algorithm \ref{alg:algorithm} only requires the calculation of $\Phi_{{X}_i;{\mathcal{X}} \rightarrow {Y}}$ once per feature. Consequently, its complexity is linear in time with respect to the number of features. Otherwise, the method becomes non-scalable and inapplicable to highly-dimensional spaces. Furthermore, we note that this Algorithm derives the size of $\mathcal{X}_*$, overcoming the length $k$ problem. Again, this is highly desirable for large feature spaces. Furthermore, in Appendix \ref{app:conv_gar}, we discuss the performance guarantees when classifying attacks using the set of features selected by our method.

\subsection{Transfer Entropy Neural Estimation}
\label{sec:measure_estimation}
In this section, we explain how we neurally estimate $\Phi_{{X}_i;{\mathcal{X}} \rightarrow {Y}}$.
Unless we restrict our domain of interest to scenarios not applicable to the real world, the calculation of $\Phi_{{X}_i;{\mathcal{X}} \rightarrow {Y}}$ will be intractable without infinite samples. For this reason, we use a neural function approximator to estimate this value. One of the most commonly estimated information theoretic quantities is MI \citep{Moon1995,Paninski2007, belghazi2018,oord2018,Poole2019}. In particular, we estimate $\Phi_{{X}_i;{\mathcal{X}} \rightarrow {Y}}$ as the difference between two $I$'s. More formally, we have:
\begin{equation}
\label{eqn:TE_from_MI}
    \Phi_{{X}_i;{\mathcal{X}} \rightarrow {Y}}  = I(\mathcal{X};Y) - I(\mathcal{X}_{\backslash X_i};Y).
\end{equation}
\textit{Derivation.} See Appendix \ref{appendix:derivation}.

The method we use for estimating MI is based on the work by Belghazi et al. \cite{belghazi2018}, as it is applicable to both continuous and discrete variables. In particular, in \cite{belghazi2018} the authors prove that the MI can be re-written via the Donsker Vardhan representation, such that: 
\begin{equation}
\label{eqn:dvrep}
    I(\mathcal{X};Y)= \sup_{T:A(\mathcal{X}) \times A(Y) \mapsto \mathbb{R}} \mathbb{E}_{P_{\mathcal{X},Y}}[T(\mathcal{X},Y)] - \log \mathbb{E}_{P_{\mathcal{X}},P_{Y}}[e^{T(\mathcal{X},Y)}]
\end{equation}
\noindent where $\mathbb{E}_{P_{\mathcal{X},Y}}$ is the expectation under the joint distribution $P_{\mathcal{X},Y}$ and $\mathbb{E}_{P_{\mathcal{X}},P_{Y}}$ is the expectation under the marginal distribution $P_{\mathcal{X}} \otimes P_{Y}$. The form of Equation \ref{eqn:dvrep} is such that it can be used for gradient ascent,  where $T(\mathcal{X},Y)$ is the output of a neural estimator, which takes as its input the full set of variables $\mathcal{X}$ and the target $Y$ sampled according to the expectation to which $T(\mathcal{X},Y)$ is subjected. For instance, $\mathbb{E}_{P_{\mathcal{X},Y}}[T(\mathcal{X},Y)]$ (the left term) is the average output of the neural function approximator when presented with $\mathcal{X}$ and $Y$ sampled jointly. Meanwhile, $\log \mathbb{E}_{P_{\mathcal{X}},P_{Y}}[e^{T(\mathcal{X},Y)}]$ (the right term) is the log of the average exponential output when  $\mathcal{X}$ and $Y$ are sampled marginally. 
The equivalence in Equation \ref{eqn:dvrep} ensures we can use the relationship in Equation \ref{eqn:TE_from_MI} to calculate $\Phi_{{X}_i;{\mathcal{X}} \rightarrow {Y}}$,  and apply Algorithm \ref{alg:algorithm} to identify our subset of features. More formally, $\Phi_{{X}_i;{\mathcal{X}} \rightarrow {Y}}$ can be written as:
\begin{equation}
\begin{split}
\label{eqn:dvrepphi}
    \Phi_{{X}_i;{\mathcal{X}} \rightarrow {Y}} & = \sup_{T:A(\mathcal{X}) \times A(Y) \mapsto \mathbb{R}} (\mathbb{E}_{P_{\mathcal{X},Y}}[T(\mathcal{X},Y)] - \log \mathbb{E}_{P_{\mathcal{X}},P_{Y}}[e^{T(\mathcal{X},Y)}] )\\ & - \sup_{U:A(\mathcal{X}_{\backslash X_i}) \times A(Y) \mapsto \mathbb{R}} (\mathbb{E}_{P_{\mathcal{X}_{\backslash X_i},Y}}[U(\mathcal{X}_{\backslash X_i},Y)] \\ & - \log \mathbb{E}_{P_{\mathcal{X}_{\backslash X_i}},P_{Y}}[e^{U(\mathcal{X}_{\backslash X_i},Y)}]).
\end{split}
\end{equation}
Calculating our measure using Equation \ref{eqn:dvrepphi} yields a real positive number, expressed in nats, which quantifies the reduction in uncertainty of the target variable due to the inclusion of feature \(X_i\). This implies that our measure - and the method employed to calculate it - offers an intuitive interpretation, aiding users not only in terms of the feature selection task but also in understanding why some features are selected over others.
\subsection{Incorporating Time Dependencies}
\label{sec:incorperating_time_dep}
We now describe the motivation of including a recurrent layer in the classification and feature selection tasks. We select an LSTM-based solution in our implementation  after comparing different architectures that integrate temporal dependencies. In Appendix \ref{app:lstm_ablation}, we report the experimental results supporting this choice.

An LSTM-based NID system receives network data of sequence size $s$, before processing it for the purpose of classification. Therefore, our NID mapping function no longer takes as its input $\xi^t$, but rather it receives a list of time ordered network data that it uses to infer the attack status. More formally $f^{LSTM}(\xi^{t-s},\xi^{t-s+1}\dots \xi^t) \in A(Y)$, where $f^{LSTM}$ is an updated mapping function such that $f^{LSTM}:A(\mathcal{X}^{t-s:t}) \mapsto A(Y)$. The set $\mathcal{X}^{t-s:t}$ in its full form is $\mathcal{X}^{t-s:t} = \{X_1^{t-s:t}, X_2^{t-s:t}... X_N^{t-s:t}\}$, and the superscript $^{t-s:t}$, as applied to variables, or sets of variables, indicates that sampling is conducted such that realizations are accompanied by the $s$ values that came before it in time, where $s$ indicates the sequence size. The recurrent layer in an LSTM has the ability to incorporate all the information within the sequence, despite the extended time horizon \cite{hochreiter1997}. This differs from standard methods for function approximation that only consider a single instance in time. Detecting dependencies in time is of importance when classifying cyberattacks. To explain why, we present the following simple example. Firstly, let us suppose a system is under attack at time $t$, it is obviously more likely to also be under attack at time $t+1$. Additionally, non-automated attacks are likely to occur at times when people are awake. Methods that account for time dependencies can exploit such relationships, boosting their detection accuracy \cite{yin2017,xu2018,naseer2018,jiang2020}. 

However, in this paper, we intend to perform the classification task on a reduced subset of selected features. Therefore, the method used to select these features should also incorporate dependencies in time (introduced before as the \textit{temporal relationship problem}). Otherwise, our feature set may omit information our classification system would typically utilize. To do so, we rewrite Equation \ref{eqn:dvrep} such that our estimation of $\Phi_{{X}_i;{\mathcal{X}} \rightarrow {Y}}$ now incorporates temporal relationships:
\begin{equation}
\begin{split}
\label{eqn:dvreptime}
    &I(\mathcal{X}^{t-s:t};Y^{t-s:t})\\& = \sup_{L:A(\mathcal{X}^{t-s:t}) \times A(Y^{t-s:t}) \mapsto \mathbb{R}} \mathbb{E}_{P_{\mathcal{X}^{t-s:t},Y^{t-s:t}}}[L(\mathcal{X}^{t-s:t},Y^{t-s:t})] \\ & -  \log \mathbb{E}_{P_{\mathcal{X}^{t-s:t}},P_{Y^{t-s:t}}}[e^{L(\mathcal{X}^{t-s:t},Y^{t-s:t})}].
\end{split}
\end{equation}

The function $L(\cdot)$ is again the output of a neural function approximator, where now the input is the variables $\mathcal{X}^{t-s:t}$ and $Y^{t-s:t}$. Each variable is sampled as chronologically ordered sequences of size $s$. These are then combined either jointly or marginally according to the expectations to which they are subjected. Furthermore, a similar update can be made to Equation \ref{eqn:dvrepphi}; although, we omit it here for space and clarity. Throughout the remainder of the paper, we will distinguish between our methodology that incorporates temporal dependencies from that which does not by using the terms, LSTM-based FSNID and FSNID, respectively.

 \subsection{Methods for Approximating State Variable Inclusion Conditions}
\label{subsection:practical}
 Since we estimate \(\Phi_{{X}_i;{\mathcal{X}} \rightarrow {Y}}\) neurally rather than calculating it directly, the values of $\Phi_{{X}_i;{\mathcal{X}} \rightarrow {Y}}$ randomly fluctuate. Therefore, the condition $\Phi_{{X}_i;{\mathcal{X}} \rightarrow {Y}} > 0$ in Algorithm \ref{alg:algorithm} is susceptible to the inclusion of uninformative variables in our target set. In this section, we explain how we approximate the condition \(\Phi_{{X}_i;{\mathcal{X}} \rightarrow {Y}} = 0\) for varying estimates. 
 
Our approach mirrors that introduced in \cite{wollstadt23}; specifically, it involves implementing a null model for comparative analysis. In this paper, we introduce a random variable \(NM\) into $\mathcal{X}$, whereas \cite{wollstadt23} remove dependencies by shuffling one of its pre-existing members. The expectation is that the ground truth should not be dependent on \(NM\), implying that \(\Phi_{NM;\mathcal{X} \rightarrow {Y}}=0\). Therefore, variables that demonstrate a significant transfer of entropy to $Y$ are expected to deviate from this null model. We assume that both the null model and the features in \(\mathcal{X}\) follow normal distributions. By adopting the Neyman-Pearson's methodology \cite{neyman_pearson_1933}, we identify variables \( {X}_i \) within \( {\mathcal{X}} \) where \(\Phi_{{X}_i;{\mathcal{X}} \rightarrow {Y}}\) has a 95\% likelihood of not conforming to the range specified by the null model. These identified variables are considered to exhibit a statistically significant divergence from the null model, thereby satisfying the condition \(\Phi_{{X}_i;{\mathcal{X}} \rightarrow {Y}} > 0\).

\section{Experimental Evaluation}
In this section, we outline the motivation and results of our experiments. Initially, we specify the research questions we aim to address in the evaluation of the method and we detail how these informed the selection of our comparative baselines. 
Our evaluation will first focus on assessing FSNID's ability to parsimoniously select features for NID. Finally, we will analyze the temporal complexity of the proposed method.

\subsection{Research Questions}
 We now outline the specific research questions that are of importance when aiming to select optimal features for NID.
 \begin{itemize}
     \item \textbf{RQ1.} \textit{To what extent does FSNID reduce the feature set compared to the baselines, and how does this affect classification performance?}
     \item \textbf{RQ2.} \textit{How does FSNID generalize across a variety of datasets with different characteristics?}
     \item \textbf{RQ3.} \textit{Is FSNID scalable in the presence of large feature spaces?}
 \end{itemize} 
\subsection{Baselines}
In this section, we introduce five methods for feature selection that we will use as comparative baselines. 

As examples of classic lightweight and linear in time methods with respect to the number of features, we selected
Permutation Importance (PI) and least absolute shrinkage and selection operator (LASSO) \cite{tibshirani1996regression,breiman2001}.  They are both highly scalable, and appropriate for the large datasets of interest. We also adopted the ultra marginal feature importance (UMFI) algorithm with optimal transport \cite{janssen2023}. We chose this technique as it is a state-of-the-art method for assigning feature importance. To the best of our knowledge, this is the only technique that attempts to natively deal with highly correlated variables, in a manner that is linear in time with respect to the number of features. This is essential in the presence of highly dimensional feature spaces. We also compare to the mutual information firefly algorithm (MIFA) \cite{selvakumar19}, as a representative and popular example of metaheuristics.
Additionally, we include the classic and widely adopted Conditional Likelihood Maximization (CLM) framework by Brown et al. \cite{brown12}. Unlike the first three baselines, the temporal complexity of MIFA and CLM is non-linear in time with respect to the number of features. Where possible, we utilize null models to determine whether or not a feature is informative. However, in the MIFA case we must derive the number of features $k$ using other methods. We explain in detail how we handle the length-$k$ problem for each baseline in Appendix \ref{app:len_k}.
 \begin{figure*}[th]%
    \subfigure{\includegraphics[width=17.5cm]{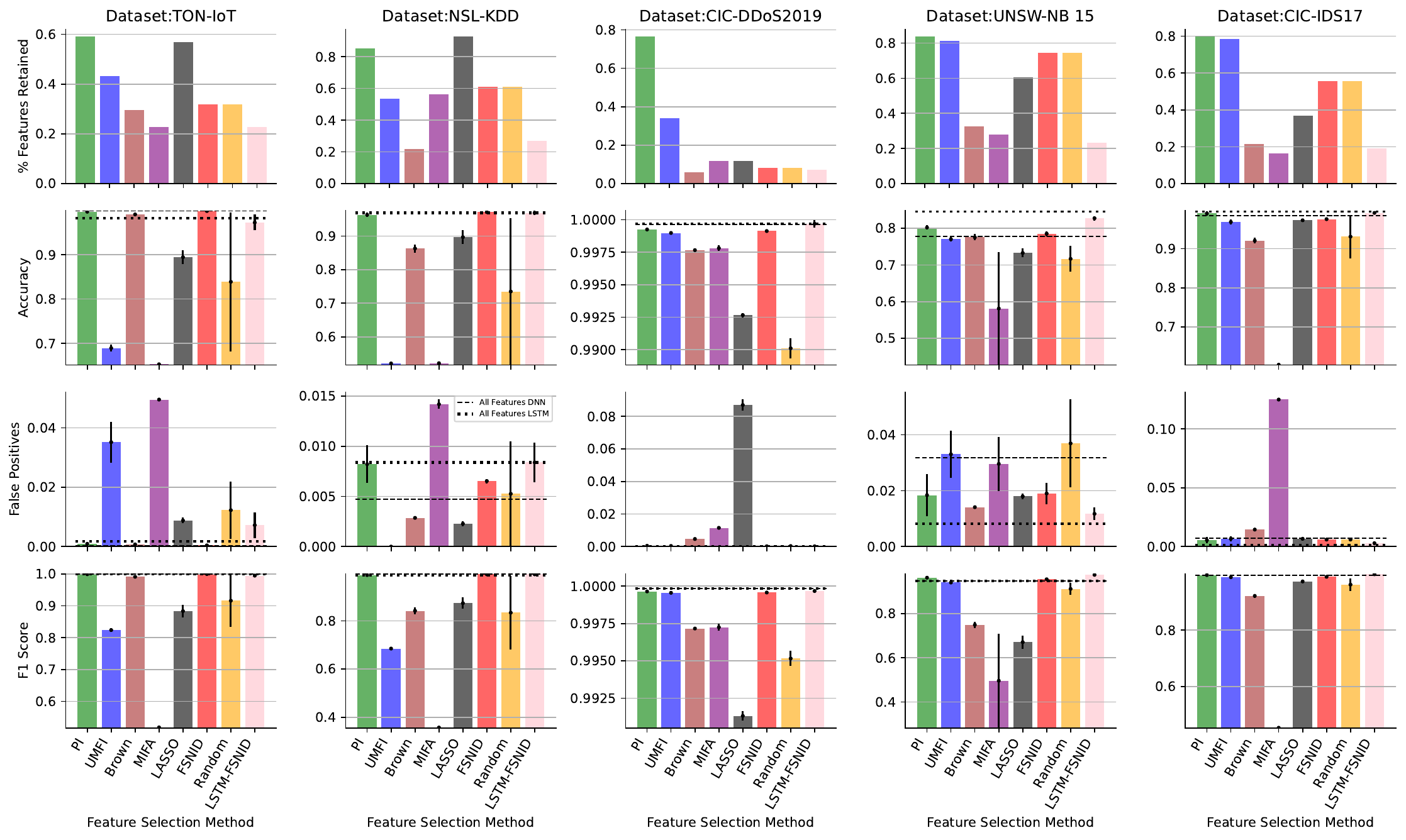} }
    \caption{Comparison of the vanilla (red bars) and LSTM-based (pink bars) versions of FSNID to PI (green bars), UMFI (blue bars), CLM (brown bars), MIFA (purple bars) and LASSO (black bars). The yellow bar corresponds to a randomly selected set of features equal in size to the set of features selected using our vanilla method. From top to Bottom we present the proportion of features that were retained using each method, the accuracy achieved during the classification task using that set, the false positive rate and F1 score.}
    \label{fig:bars}
\end{figure*}
\begin{figure*}[th]%
    \subfigure{\includegraphics[width=17.5cm]{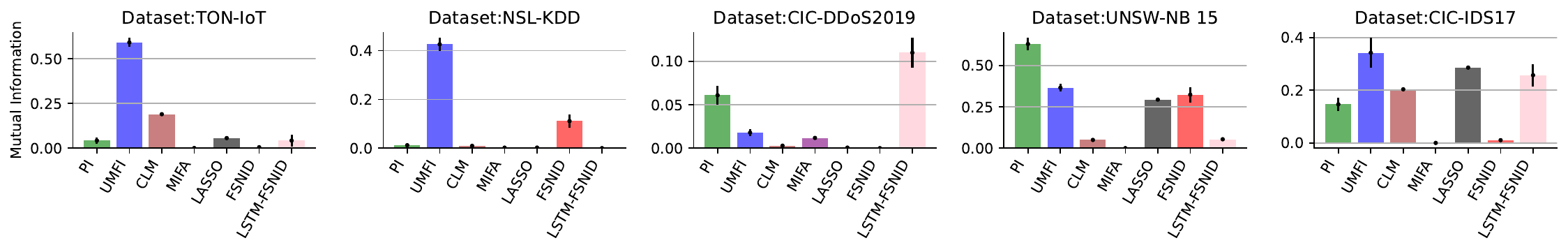} }
    \caption{Performance comparison of FSNID against the chosen baselines in terms of their ability to deal with highly correlated features. Specifically, we plot the average MI shared between the top three features for each method with respect to each dataset.}
    \label{fig:mibars}
\end{figure*}

\begin{figure}[th]%
    \subfigure{\includegraphics[width=0.8\columnwidth]{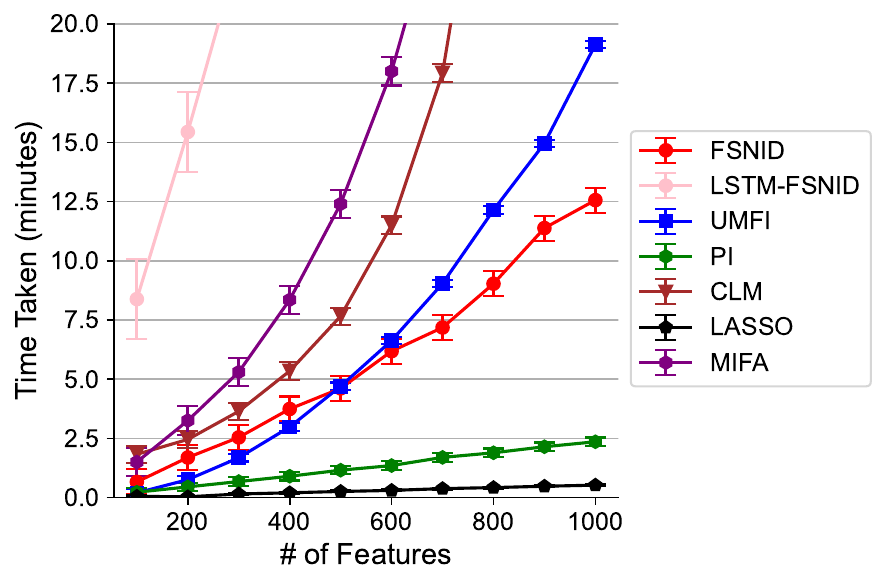} }
    \caption{Temporal complexity of FSNID and comparators with respect to the number of features. }
    \label{fig:comp_analysis}
\end{figure}

 \subsection{Feature Selection Performance Evaluation}
 We first address \textbf{RQ1} and \textbf{RQ2} by presenting the results of an extensive evaluation, comparing the feature selection performance of FSNID and the baselines.

\subsubsection{Experimental Settings} First, we introduce the datasets used for the evaluation.  We use the train-test split datasets if provided. Otherwise, we apply a standard $80:20$ split. For a more detailed dataset description with specific statistics please refer to Appendix \ref{app:dataset}. For this study, we use datasets that explore a range of network and attck types. 

The datasets used in this study provide a comprehensive view of network traffic and differ significantly in their attributes and focus. The TON-IoT dataset offers diverse IoT and IIoT sensor data combined with system traces from Linux and Windows hosts, capturing various system activities and featuring multiple attack types and benign data \cite{moustafa2021}. The NSL-KDD dataset, addressing the shortcomings of the KDD'99 dataset, includes numeric, binary, and nominal features, categorizing attacks into DoS, R2L, U2R, and Probe, with a variety of attack styles applicable to non-IoT networks \cite{tavallaee2009,selvakumar19}. The CIC-DDoS2019 dataset focuses on DDoS attacks and real-world traffic conditions, particularly highlighting the unique UDPLag attacks \cite{sharafaldin2019}. The UNSW-NB 15 dataset, consisting of multiple features and attack groups, requires consideration of time dependencies for high classification accuracy. Finally, the CICIDS17 dataset captures benign traffic and state-of-the-art attacks across various network flow features, ensuring real-world conditions are mirrored \cite{sharafaldin2018}. 

These datasets essentially encompass a broad spectrum of attacks, each varying in frequency and occurring across different applications, systems, or devices. As a result, by evaluating this combination of datasets we aim to answer \textbf{RQ2}. In Figure \ref{fig:bars}, we present the classification accuracy achieved when using all features compared to the case in which only subsets (selected according to the methods taken into consideration) are used. Furthermore, we also present false positives and F1 scores. These are of key importance when assessing NID due to the presence of imbalanced datasets (e.g., CIC-DDoS2019). Therefore Figure \ref{fig:bars}, allows direct comparison of each methods ability to select a useful feature set. To verify these methods outperform a stochastic baseline, we compare the results to a set of randomly chosen features, equal in size to the set selected by our vanilla method. 
 Additionally, in Figure \ref{fig:mibars}, we examine the MI between the top three features. This analysis shows how each method addresses the \textit{redundancy problem}.
The calculation of $\Phi_{{X}_i;{\mathcal{X}}\rightarrow {Y}}$ and the classification results are obtained over 5 runs. We adopt $\pm 95 \%$ confidence intervals for the classification task. 
Hyperparameter tuning is conducted via a simple grid search (for a full discussion, please refer to Appendix \ref{appendix:hyper}). Finally, in Appendix \ref{app:Botiot_expts} we present further experimental results obtained using the Bot-IoT dataset. 
\subsubsection{Experimental Results}\label{sec:fs_expts}  We now introduce and discuss the results achieved for each dataset.

\noindent \textbf{TON-IoT.}
In this dataset, FSNID achieves near-optimal performance despite a large reduction in the feature space; it is apparent that UMFI's performance does not surpass that of a randomly chosen set of variables equivalent in size to our DNN-based approach. Notably, this occurs even though UMFI selects a larger number of features than FSNID. This can be explained with reference to Figure \ref{fig:mibars}, in which it is possible to observe higher correlation among the features chosen by UMFI. This suggests that many of the attributes selected by UMFI might be superfluous for effective classification. Such redundant or highly correlated features can impede a model's ability to detect attacks. The MIFA method correctly evaluates the formed feature subsets (using MI). However, the heuristic employed to update these subsets is inefficient. This approach assigns uniform probabilities across all features with the potential to be switched on or off, without considering their individual contributions to overall performance. Consequently, to discover a viable feature set using MIFA, one must either evaluate a very large number of feature subsets (fireflies) or allow a significant number of feature-flipping events. Both options increase the computational requirements beyond practical limits for large datasets. The LASSO algorithm selects a large number of features without achieving competitive performance. LASSO exclusively considers linear relationships; for this reason, it is unable to incorporate synergistic interactions or resolve complex redundancies. The inability to detect synergistic relationships can lead to too few features being detected, while the inability to handle highly correlated features can lead to over-selection \cite{friedman2010regularization}. In this case, the latter occurs. While CLM selects fewer features, it is marginally outperformed by FSNID.

\noindent \textbf{NSL-KDD.} We observe that the performance of UMFI and LASSO is poor, given the amount of features they select. This is likely due to over-representation of correlated features, which is clearly evidenced for UMFI in Figure \ref{fig:mibars}. In this case, we observe that CLM selects the fewest number of features, although this does prevent the model learning an optimal detection strategy. This underperformance is primarily due to an inability to recognize synergistic features as important. To clarify, CLM focuses on adding individual features based on their ability to increase MI, neglecting the potential benefits of their combinations: as a result, synergistic combinations remain undiscovered. Overlooking such interactions consistently leads to too few features being selected. This has an impact on performance. On the other hand, FSNID natively considers synergistic interactions, leading to the selection of an appropriate feature set which in turn leads to good performance. 

\noindent \textbf{CIC-DDoS2019.}
For this dataset, we observe that FSNID achieves comparable performance with both PI and UMFI despite these methods selecting much larger feature sets. CLM's inability to consider purely synergistic interactions ensures that it again selects the fewest number of features. Despite this, it outperforms both LASSO and MIFA.

\noindent \textbf{UNSW-NB 15.}
In this scenario, it is observed that although the LSTMFSNID method selects a limited number of features, it performs comparably with a full feature subset. Given the UNSW-NB 15 dataset is characterized by attacks that are time-dependent, the use of an LSTM not only reduces the required number of features but also leads to improved performance. CLM here selects the fewest features and performs well. However, this methods inability to solve the complexity problem means the selection of this set took 2.5x the amount of time needed for FSNID, as presented in Appendix \ref{app:runtimes}. Meanwhile, the MIFA algorithm performs worse than random selection, although this is likely due to the low number of features MIFA selected. 

\noindent \textbf{CIC-IDS 17.}
Also in this case, we observe that FSNID drastically reduces the size of the set of features, while not significantly affecting the predictive power of the model. LSTM-based FSNID is able to achieve even better performance, for similar reasons to that described for UNSW-NB 15.
To summarize, in these experiments, by means of a variety of datasets, we have demonstrated our method's ability to select a reduced set of highly informative features that result in near-optimal classification accuracy. In other words, we have provided empirical evidence to address both \textbf{RQ1} and \textbf{RQ2}. Furthermore, we have demonstrated that by exploiting the temporal relationships present in the data, we can enhance classification performance while using even smaller sets of features. 
 
\subsection{Analysis of the Computational Cost}\label{sec:comp_expts}
In this section, we present a set of results that provides evidence for addressing \textbf{RQ3} following an experimental methodology similar to that presented in \cite{janssen2023}.  
\subsubsection{Experimental Settings} In the following experiments, we intend to demonstrate that the temporal complexity of our method scales linearly with respect to the number of features. We are not interested, at this stage, in each methods ability to select features. Consequently, to reduce the computation associated with this evaluation, we use simple synthetic data. The synthetic features ($X_i \in \mathcal{X}$) and the target ($Y$) are independent stochastic binary arrays of length 500. Meanwhile, for a discussion of the hyperparameter selection we refer to Appendix \ref{appendix:hyper}. 
\subsubsection{Experimental Results} The results indicate that the complexity of our method scales approximately linearly in time with respect to the number of features. Although UMFI is theoretically linear in time, its practical implementation involves a pre-processing step that leads to the curve observed in Figure \ref{fig:comp_analysis}. 
CLM exhibits a non-linear time complexity with respect to the total number of features $N$ and the selected number $k$, scaling proportionally to $\mathcal{O}(Nk-\frac{k(k-1)}{2})$. This complexity arises as each selected feature must maximally increase the correlation with the target, making this a greedy maximization algorithm, where this complexity is a common result. On the other hand, the firefly algorithm scales proportionally with respect to the number of features, but is quadratic in time with respect to the number of fireflies (potential feature subsets). In practice, it is recommended to increase the number of fireflies as the number of features increases. For our experiments, we chose the number of fireflies to be equal to half the number of features, yielding the results presented in Figure \ref{fig:comp_analysis}. Overall, this dependence on both the features and fireflies makes the algorithm exceedingly complex. These aspects render our method more suitable for high-dimensional feature spaces than any of the baselines discussed thus far. However, the scalability of both LASSO and PI exceeds that of any other baselines, despite their troubles selecting features. Our results also highlight the additional computational load required to integrate temporal dependencies. 
In any case, the LSTM-based version of FSNID still shows a linear time complexity with respect to to the number of features. 
\section{Conclusion}
Feature selection is a fundamental problem in network intrusion detection. This is due to the computational cost associated to large input spaces and the requirement of minimizing monitoring for both economic and strategic reasons.
In this paper, we have introduced FSNID, a new information-theoretic framework designed to select features for detecting network intrusions. Our approach estimates a TE-based `feature importance' using a neural function approximator. FSNID effectively identifies the minimal number of features even if they are highly correlated, while its computation time scales linearly relative to the number of features. All of these are desirable attributes when applied to highly-dimensional network data. The estimation of the measure central to FSNID is agnostic to the function approximation technique used. By integrating a recurrent layer into the neural architecture, it is possible to incorporate time-dependent information into our feature importance calculation. This enables a further reduction in the number of features selected when compared to our vanilla method. In our experimental evaluation, we have shown that FSNID can consistently identify a lightweight set of network features, without notably compromising classification performance.

\bibliographystyle{ACM-Reference-Format}
\bibliography{sample-base}


\begin{thebibliography}{73}


\ifx \showCODEN    \undefined \def \showCODEN     #1{\unskip}     \fi
\ifx \showDOI      \undefined \def \showDOI       #1{#1}\fi
\ifx \showISBNx    \undefined \def \showISBNx     #1{\unskip}     \fi
\ifx \showISBNxiii \undefined \def \showISBNxiii  #1{\unskip}     \fi
\ifx \showISSN     \undefined \def \showISSN      #1{\unskip}     \fi
\ifx \showLCCN     \undefined \def \showLCCN      #1{\unskip}     \fi
\ifx \shownote     \undefined \def \shownote      #1{#1}          \fi
\ifx \showarticletitle \undefined \def \showarticletitle #1{#1}   \fi
\ifx \showURL      \undefined \def \showURL       {\relax}        \fi
\providecommand\bibfield[2]{#2}
\providecommand\bibinfo[2]{#2}
\providecommand\natexlab[1]{#1}
\providecommand\showeprint[2][]{arXiv:#2}

\bibitem[Ahmed et~al\mbox{.}(2016)]%
        {ahmed2016}
\bibfield{author}{\bibinfo{person}{Mohiuddin Ahmed},
  \bibinfo{person}{Abdun~Naser Mahmood}, {and} \bibinfo{person}{Jiankun Hu}.}
  \bibinfo{year}{2016}\natexlab{}.
\newblock \showarticletitle{A survey of network anomaly detection techniques}.
\newblock \bibinfo{journal}{\emph{Journal of Network and Computer
  Applications}} \bibinfo{volume}{60}, \bibinfo{number}{3}
  (\bibinfo{year}{2016}), \bibinfo{pages}{19--31}.
\newblock


\bibitem[Ahmim et~al\mbox{.}(2019)]%
        {ahmim2019}
\bibfield{author}{\bibinfo{person}{Ahmed Ahmim}, \bibinfo{person}{Leandros
  Maglaras}, \bibinfo{person}{Mohamed~Amine Ferrag}, \bibinfo{person}{Makhlouf
  Derdour}, {and} \bibinfo{person}{Helge Janicke}.}
  \bibinfo{year}{2019}\natexlab{}.
\newblock \showarticletitle{A novel hierarchical intrusion detection system
  based on decision tree and rules-based models}. In
  \bibinfo{booktitle}{\emph{19th International Conference on Distributed
  Computing in Sensor Systems (DCOSS'19)}}. \bibinfo{publisher}{IEEE},
  \bibinfo{pages}{228--233}.
\newblock


\bibitem[Amiri et~al\mbox{.}(2011)]%
        {amiri11}
\bibfield{author}{\bibinfo{person}{Fatemeh Amiri},
  \bibinfo{person}{MohammadMahdi~Rezaei Yousefi}, \bibinfo{person}{Caro Lucas},
  \bibinfo{person}{Azadeh Shakery}, {and} \bibinfo{person}{Nasser Yazdani}.}
  \bibinfo{year}{2011}\natexlab{}.
\newblock \showarticletitle{Mutual information-based feature selection for
  intrusion detection systems}.
\newblock \bibinfo{journal}{\emph{Journal of Network and Computer
  Applications}} \bibinfo{volume}{34}, \bibinfo{number}{4}
  (\bibinfo{year}{2011}), \bibinfo{pages}{1184--1199}.
\newblock


\bibitem[Battiti(1994)]%
        {battiti94}
\bibfield{author}{\bibinfo{person}{R. Battiti}.}
  \bibinfo{year}{1994}\natexlab{}.
\newblock \showarticletitle{Using mutual information for selecting features in
  supervised neural net learning}.
\newblock \bibinfo{journal}{\emph{IEEE Transactions on Neural Networks}}
  \bibinfo{volume}{5}, \bibinfo{number}{4} (\bibinfo{year}{1994}),
  \bibinfo{pages}{537--550}.
\newblock


\bibitem[Belghazi et~al\mbox{.}(2018)]%
        {belghazi2018}
\bibfield{author}{\bibinfo{person}{Ishmael Belghazi}, \bibinfo{person}{Sai
  Rajeswar}, \bibinfo{person}{Aristide Baratin}, \bibinfo{person}{R~Devon
  Hjelm}, {and} \bibinfo{person}{Aaron Courville}.}
  \bibinfo{year}{2018}\natexlab{}.
\newblock \showarticletitle{MINE: Mutual Information Neural Estimation}. In
  \bibinfo{booktitle}{\emph{19th International Conference on Machine Learning
  (ICML'18)}}. \bibinfo{publisher}{PMLR}, \bibinfo{pages}{531--540}.
\newblock


\bibitem[Bostani and Sheikhan(2017)]%
        {bostani17}
\bibfield{author}{\bibinfo{person}{Hamid Bostani} {and}
  \bibinfo{person}{Mansour Sheikhan}.} \bibinfo{year}{2017}\natexlab{}.
\newblock \showarticletitle{Hybrid of binary gravitational search algorithm and
  mutual information for feature selection in intrusion detection systems}.
\newblock \bibinfo{journal}{\emph{Soft Computing}}  \bibinfo{volume}{21}
  (\bibinfo{year}{2017}), \bibinfo{pages}{2307--2324}.
\newblock


\bibitem[Breiman(2001)]%
        {breiman2001}
\bibfield{author}{\bibinfo{person}{Leo Breiman}.}
  \bibinfo{year}{2001}\natexlab{}.
\newblock \showarticletitle{Random forests}.
\newblock \bibinfo{journal}{\emph{Machine Learning}} \bibinfo{volume}{45},
  \bibinfo{number}{1} (\bibinfo{year}{2001}), \bibinfo{pages}{5--32}.
\newblock


\bibitem[Brown et~al\mbox{.}(2012)]%
        {brown12}
\bibfield{author}{\bibinfo{person}{Gavin Brown}, \bibinfo{person}{Adam Pocock},
  \bibinfo{person}{Ming-Jie Zhao}, {and} \bibinfo{person}{Mikel Luj{{\'a}}n}.}
  \bibinfo{year}{2012}\natexlab{}.
\newblock \showarticletitle{Conditional Likelihood Maximisation: A Unifying
  Framework for Information Theoretic Feature Selection}.
\newblock \bibinfo{journal}{\emph{Journal of Machine Learning Research}}
  \bibinfo{volume}{13}, \bibinfo{number}{2} (\bibinfo{year}{2012}),
  \bibinfo{pages}{27--66}.
\newblock


\bibitem[Catav et~al\mbox{.}(2021)]%
        {catav21}
\bibfield{author}{\bibinfo{person}{Amnon Catav}, \bibinfo{person}{Boyang Fu},
  \bibinfo{person}{Yazeed Zoabi}, \bibinfo{person}{Ahuva Libi~Weiss Meilik},
  \bibinfo{person}{Noam Shomron}, \bibinfo{person}{Jason Ernst},
  \bibinfo{person}{Sriram Sankararaman}, {and} \bibinfo{person}{Ran
  Gilad-Bachrach}.} \bibinfo{year}{2021}\natexlab{}.
\newblock \showarticletitle{Marginal Contribution Feature Importance - an
  Axiomatic Approach for Explaining Data}. In \bibinfo{booktitle}{\emph{19th
  International Conference on Machine Learning (ICML'21)}}.
  \bibinfo{publisher}{PMLR}, \bibinfo{pages}{1324--1335}.
\newblock


\bibitem[Chen et~al\mbox{.}(2023)]%
        {chen23}
\bibfield{author}{\bibinfo{person}{Jinfu Chen}, \bibinfo{person}{Yuhao Chen},
  \bibinfo{person}{Saihua Cai}, \bibinfo{person}{Shang Yin},
  \bibinfo{person}{Lingling Zhao}, {and} \bibinfo{person}{Zikang Zhang}.}
  \bibinfo{year}{2023}\natexlab{}.
\newblock \showarticletitle{An optimized feature extraction algorithm for
  abnormal network traffic detection}.
\newblock \bibinfo{journal}{\emph{Future Generation Computer Systems}}
  \bibinfo{volume}{149}, \bibinfo{number}{1} (\bibinfo{year}{2023}),
  \bibinfo{pages}{330--342}.
\newblock


\bibitem[Chen et~al\mbox{.}(2018)]%
        {chen18}
\bibfield{author}{\bibinfo{person}{Jianbo Chen}, \bibinfo{person}{Le Song},
  \bibinfo{person}{Martin Wainwright}, {and} \bibinfo{person}{Michael Jordan}.}
  \bibinfo{year}{2018}\natexlab{}.
\newblock \showarticletitle{Learning to explain: An information-theoretic
  perspective on model interpretation}. In \bibinfo{booktitle}{\emph{19th
  International Conference on Machine Learning (ICML'18)}}.
  \bibinfo{publisher}{PMLR}, \bibinfo{pages}{883--892}.
\newblock


\bibitem[Chou et~al\mbox{.}(2008)]%
        {chou2008}
\bibfield{author}{\bibinfo{person}{Te-Shun Chou}, \bibinfo{person}{Kang~K Yen},
  {and} \bibinfo{person}{Jun Luo}.} \bibinfo{year}{2008}\natexlab{}.
\newblock \showarticletitle{Network intrusion detection design using feature
  selection of soft computing paradigms}.
\newblock \bibinfo{journal}{\emph{International Journal of Computer and
  Information Engineering}} \bibinfo{volume}{2}, \bibinfo{number}{11}
  (\bibinfo{year}{2008}), \bibinfo{pages}{3722--3734}.
\newblock


\bibitem[Covert et~al\mbox{.}(2020)]%
        {covert2020}
\bibfield{author}{\bibinfo{person}{Ian~C. Covert}, \bibinfo{person}{Scott
  Lundberg}, {and} \bibinfo{person}{Su-In Lee}.}
  \bibinfo{year}{2020}\natexlab{}.
\newblock \showarticletitle{Understanding Global Feature Contributions with
  Additive Importance Measures}. In \bibinfo{booktitle}{\emph{34th Annual
  Conference on Neural Information Processing Systems (NeurIPS'20)}}.
  \bibinfo{pages}{17212--17223}.
\newblock


\bibitem[Covert et~al\mbox{.}(2023)]%
        {covert23}
\bibfield{author}{\bibinfo{person}{Ian~Connick Covert}, \bibinfo{person}{Wei
  Qiu}, \bibinfo{person}{Mingyu Lu}, \bibinfo{person}{Na~Yoon Kim},
  \bibinfo{person}{Nathan~J White}, {and} \bibinfo{person}{Su-In Lee}.}
  \bibinfo{year}{2023}\natexlab{}.
\newblock \showarticletitle{Learning to Maximize Mutual Information for Dynamic
  Feature Selection}. In \bibinfo{booktitle}{\emph{40th International
  Conference on Machine Learning (ICML'23)}}. \bibinfo{publisher}{PMLR},
  \bibinfo{pages}{6424--6447}.
\newblock


\bibitem[Debeer and Strobl(2020)]%
        {debeer20}
\bibfield{author}{\bibinfo{person}{Dries Debeer} {and} \bibinfo{person}{Carolin
  Strobl}.} \bibinfo{year}{2020}\natexlab{}.
\newblock \showarticletitle{Conditional permutation importance revisited}.
\newblock \bibinfo{journal}{\emph{BMC Bioinformatics}} \bibinfo{volume}{21},
  \bibinfo{number}{1} (\bibinfo{date}{07} \bibinfo{year}{2020}),
  \bibinfo{pages}{1--30}.
\newblock


\bibitem[Denning(1987)]%
        {denning87}
\bibfield{author}{\bibinfo{person}{Dorothy Denning}.}
  \bibinfo{year}{1987}\natexlab{}.
\newblock \showarticletitle{Algorithmic Enumeration of Ideal Classes for
  Quaternion Orders}.
\newblock \bibinfo{journal}{\emph{IEEE Transactions on Software Engineering}}
  \bibinfo{volume}{12}, \bibinfo{number}{2} (\bibinfo{year}{1987}),
  \bibinfo{pages}{222--232}.
\newblock


\bibitem[Dobigeon et~al\mbox{.}(2007)]%
        {dobigeon07}
\bibfield{author}{\bibinfo{person}{Nicolas Dobigeon},
  \bibinfo{person}{Jean-Yves Tourneret}, {and} \bibinfo{person}{Manuel Davy}.}
  \bibinfo{year}{2007}\natexlab{}.
\newblock \showarticletitle{Joint segmentation of piecewise constant
  autoregressive processes by using a hierarchical model and a Bayesian
  sampling approach}.
\newblock \bibinfo{journal}{\emph{IEEE Transactions on Signal Processing}}
  \bibinfo{volume}{55}, \bibinfo{number}{4} (\bibinfo{year}{2007}),
  \bibinfo{pages}{1251--1263}.
\newblock


\bibitem[Friedman et~al\mbox{.}(2010)]%
        {friedman2010regularization}
\bibfield{author}{\bibinfo{person}{Jerome Friedman}, \bibinfo{person}{Trevor
  Hastie}, {and} \bibinfo{person}{Rob Tibshirani}.}
  \bibinfo{year}{2010}\natexlab{}.
\newblock \showarticletitle{Regularization paths for generalized linear models
  via coordinate descent}.
\newblock \bibinfo{journal}{\emph{Journal of statistical software}}
  \bibinfo{volume}{33}, \bibinfo{number}{1} (\bibinfo{year}{2010}),
  \bibinfo{pages}{1}.
\newblock


\bibitem[Frye et~al\mbox{.}(2020)]%
        {Frye2020}
\bibfield{author}{\bibinfo{person}{Christopher Frye}, \bibinfo{person}{Colin
  Rowat}, {and} \bibinfo{person}{Ilya Feige}.} \bibinfo{year}{2020}\natexlab{}.
\newblock \showarticletitle{Asymmetric Shapley Values: Incorporating Causal
  Knowledge into Model-Agnostic Explainability}. In
  \bibinfo{booktitle}{\emph{34th Annual Conference on Neural Information
  Processing Systems (NeurIPS'20)}}. \bibinfo{pages}{1229--1239}.
\newblock


\bibitem[Gao et~al\mbox{.}(2016)]%
        {gao16}
\bibfield{author}{\bibinfo{person}{Shuyang Gao}, \bibinfo{person}{Greg
  Ver~Steeg}, {and} \bibinfo{person}{Aram Galstyan}.}
  \bibinfo{year}{2016}\natexlab{}.
\newblock \showarticletitle{Variational Information Maximization for Feature
  Selection}. In \bibinfo{booktitle}{\emph{30th Annual Conference on Neural
  Information Processing Sytems (NeurIPS'16)}},
  \bibfield{editor}{\bibinfo{person}{D.~Lee}, \bibinfo{person}{M.~Sugiyama},
  \bibinfo{person}{U.~Luxburg}, \bibinfo{person}{I.~Guyon}, {and}
  \bibinfo{person}{R.~Garnett}} (Eds.). \bibinfo{pages}{1285--1312}.
\newblock


\bibitem[Gregorutti et~al\mbox{.}(2017)]%
        {gregorutti2017correlation}
\bibfield{author}{\bibinfo{person}{Baptiste Gregorutti},
  \bibinfo{person}{Bertrand Michel}, {and} \bibinfo{person}{Philippe
  Saint-Pierre}.} \bibinfo{year}{2017}\natexlab{}.
\newblock \showarticletitle{Correlation and variable importance in random
  forests}.
\newblock \bibinfo{journal}{\emph{Statistics and Computing}}
  \bibinfo{volume}{27} (\bibinfo{year}{2017}), \bibinfo{pages}{659--678}.
\newblock


\bibitem[Harle et~al\mbox{.}(2014)]%
        {harle14}
\bibfield{author}{\bibinfo{person}{Flore Harle}, \bibinfo{person}{Florent
  Chatelain}, \bibinfo{person}{Cedric Gouy-Pailler}, {and}
  \bibinfo{person}{Sophie Achard}.} \bibinfo{year}{2014}\natexlab{}.
\newblock \showarticletitle{Rank-based multiple change-point detection in
  multivariate time series}. In \bibinfo{booktitle}{\emph{2014 22nd European
  Signal Processing Conference (EUSIPCO'14)}}. \bibinfo{publisher}{IEEE},
  \bibinfo{pages}{1337--1341}.
\newblock


\bibitem[Hochreiter and Schmidhuber(1997)]%
        {hochreiter1997}
\bibfield{author}{\bibinfo{person}{Sepp Hochreiter} {and}
  \bibinfo{person}{J{\"u}rgen Schmidhuber}.} \bibinfo{year}{1997}\natexlab{}.
\newblock \showarticletitle{Long short-term memory}.
\newblock \bibinfo{journal}{\emph{Neural Computation}} \bibinfo{volume}{9},
  \bibinfo{number}{8} (\bibinfo{year}{1997}), \bibinfo{pages}{1735--1780}.
\newblock


\bibitem[{IBM}(2023)]%
        {ibmreport2023}
\bibfield{author}{\bibinfo{person}{{IBM}}.} \bibinfo{year}{2023}\natexlab{}.
\newblock \bibinfo{title}{{Cost of a Data Breach Report 2023}}.
\newblock
\newblock
\urldef\tempurl%
\url{https://www.ibm.com/reports/data-breach}
\showURL{%
\tempurl}


\bibitem[Jabez and Muthukumar(2015)]%
        {jabez15}
\bibfield{author}{\bibinfo{person}{Ja Jabez} {and} \bibinfo{person}{B
  Muthukumar}.} \bibinfo{year}{2015}\natexlab{}.
\newblock \showarticletitle{Intrusion Detection System (IDS): Anomaly detection
  using outlier detection approach}.
\newblock \bibinfo{journal}{\emph{Procedia Computer Science}}
  \bibinfo{volume}{48} (\bibinfo{year}{2015}), \bibinfo{pages}{338--346}.
\newblock


\bibitem[Janssen et~al\mbox{.}(2023)]%
        {janssen2023}
\bibfield{author}{\bibinfo{person}{Joseph Janssen}, \bibinfo{person}{Vincent
  Guan}, {and} \bibinfo{person}{Elina Robeva}.}
  \bibinfo{year}{2023}\natexlab{}.
\newblock \showarticletitle{Ultra-marginal Feature Importance: Learning from
  Data with Causal Guarantees}. In \bibinfo{booktitle}{\emph{26th International
  Conference on Artificial Intelligence and Statistics (AISTATS'23)}}.
  \bibinfo{publisher}{PMLR}, \bibinfo{pages}{10782--10814}.
\newblock


\bibitem[Janzing et~al\mbox{.}(2020)]%
        {janzing20}
\bibfield{author}{\bibinfo{person}{Dominik Janzing}, \bibinfo{person}{Lenon
  Minorics}, {and} \bibinfo{person}{Patrick Bloebaum}.}
  \bibinfo{year}{2020}\natexlab{}.
\newblock \showarticletitle{Feature relevance quantification in explainable AI:
  A causal problem}. In \bibinfo{booktitle}{\emph{23rd International Conference
  of Artificial Intelligence and Statistics (AISTATS'20)}}.
  \bibinfo{publisher}{PMLR}, \bibinfo{pages}{2907--2916}.
\newblock


\bibitem[Javaid et~al\mbox{.}(2016)]%
        {javaid2016}
\bibfield{author}{\bibinfo{person}{Ahmad Javaid}, \bibinfo{person}{Quamar
  Niyaz}, \bibinfo{person}{Weiqing Sun}, {and} \bibinfo{person}{Mansoor Alam}.}
  \bibinfo{year}{2016}\natexlab{}.
\newblock \showarticletitle{A deep learning approach for network intrusion
  detection system}. \bibinfo{publisher}{EAI}, \bibinfo{pages}{21--26}.
\newblock


\bibitem[Jia et~al\mbox{.}(2019)]%
        {jia2019}
\bibfield{author}{\bibinfo{person}{Yang Jia}, \bibinfo{person}{Meng Wang},
  {and} \bibinfo{person}{Yagang Wang}.} \bibinfo{year}{2019}\natexlab{}.
\newblock \showarticletitle{Network intrusion detection algorithm based on deep
  neural network}.
\newblock \bibinfo{journal}{\emph{IET Information Security}}
  \bibinfo{volume}{13}, \bibinfo{number}{1} (\bibinfo{year}{2019}),
  \bibinfo{pages}{48--53}.
\newblock


\bibitem[Jiang et~al\mbox{.}(2020)]%
        {jiang2020}
\bibfield{author}{\bibinfo{person}{Kaiyuan Jiang}, \bibinfo{person}{Wenya
  Wang}, \bibinfo{person}{Aili Wang}, {and} \bibinfo{person}{Haibin Wu}.}
  \bibinfo{year}{2020}\natexlab{}.
\newblock \showarticletitle{Network intrusion detection combined hybrid
  sampling with deep hierarchical network}.
\newblock \bibinfo{journal}{\emph{IEEE Access}}  \bibinfo{volume}{8}
  (\bibinfo{year}{2020}), \bibinfo{pages}{32464--32476}.
\newblock


\bibitem[Karatas et~al\mbox{.}(2020)]%
        {karatas20}
\bibfield{author}{\bibinfo{person}{Gozde Karatas}, \bibinfo{person}{Onder
  Demir}, {and} \bibinfo{person}{Ozgur~Koray Sahingoz}.}
  \bibinfo{year}{2020}\natexlab{}.
\newblock \showarticletitle{Increasing the Performance of Machine
  Learning-Based IDSs on an Imbalanced and Up-to-Date Dataset}.
\newblock \bibinfo{journal}{\emph{IEEE Access}}  \bibinfo{volume}{8}
  (\bibinfo{year}{2020}), \bibinfo{pages}{32150--32162}.
\newblock


\bibitem[Kevric et~al\mbox{.}(2017)]%
        {kevric2017}
\bibfield{author}{\bibinfo{person}{Jasmin Kevric}, \bibinfo{person}{Samed
  Jukic}, {and} \bibinfo{person}{Abdulhamit Subasi}.}
  \bibinfo{year}{2017}\natexlab{}.
\newblock \showarticletitle{An effective combining classifier approach using
  tree algorithms for network intrusion detection}.
\newblock \bibinfo{journal}{\emph{Neural Computing and Applications}}
  \bibinfo{volume}{28}, \bibinfo{number}{1} (\bibinfo{year}{2017}),
  \bibinfo{pages}{1051--1058}.
\newblock


\bibitem[Khammassi and Krichen(2017)]%
        {khammassi17}
\bibfield{author}{\bibinfo{person}{Chaouki Khammassi} {and}
  \bibinfo{person}{Saoussen Krichen}.} \bibinfo{year}{2017}\natexlab{}.
\newblock \showarticletitle{A GA-LR wrapper approach for feature selection in
  network intrusion detection}.
\newblock \bibinfo{journal}{\emph{Computers \& Security}}  \bibinfo{volume}{70}
  (\bibinfo{year}{2017}), \bibinfo{pages}{255--277}.
\newblock


\bibitem[Khammassi and Krichen(2020)]%
        {khammassi20}
\bibfield{author}{\bibinfo{person}{Chaouki Khammassi} {and}
  \bibinfo{person}{Saoussen Krichen}.} \bibinfo{year}{2020}\natexlab{}.
\newblock \showarticletitle{A NSGA2-LR wrapper approach for feature selection
  in network intrusion detection}.
\newblock \bibinfo{journal}{\emph{Computer Networks}}  \bibinfo{volume}{172}
  (\bibinfo{year}{2020}), \bibinfo{pages}{107183}.
\newblock


\bibitem[Koroniotis et~al\mbox{.}(2019)]%
        {koroniotis2019}
\bibfield{author}{\bibinfo{person}{Nickolaos Koroniotis}, \bibinfo{person}{Nour
  Moustafa}, \bibinfo{person}{Elena Sitnikova}, {and} \bibinfo{person}{Benjamin
  Turnbull}.} \bibinfo{year}{2019}\natexlab{}.
\newblock \showarticletitle{{Towards the development of realistic botnet
  dataset in the internet of things for network forensic analytics: Bot-IoT
  dataset}}.
\newblock \bibinfo{journal}{\emph{Future Generation Computer Systems}}
  \bibinfo{volume}{100} (\bibinfo{year}{2019}), \bibinfo{pages}{779--796}.
\newblock


\bibitem[Kumar et~al\mbox{.}(2020)]%
        {Kumar2020}
\bibfield{author}{\bibinfo{person}{Elizabeth Kumar}, \bibinfo{person}{Suresh
  Venkatasubramanian}, \bibinfo{person}{Carlos Scheidegger}, {and}
  \bibinfo{person}{Sorelle~A. Friedler}.} \bibinfo{year}{2020}\natexlab{}.
\newblock \showarticletitle{Problems with Shapley-Value-Based Explanations as
  Feature Importance Measures}. In \bibinfo{booktitle}{\emph{37th International
  Conference on Machine Learning (ICML'20)}}. \bibinfo{publisher}{PMLR},
  \bibinfo{pages}{5491--5500}.
\newblock


\bibitem[Lachenbruch and Goldstein(1979)]%
        {lachenbruch1979}
\bibfield{author}{\bibinfo{person}{Peter~A Lachenbruch} {and}
  \bibinfo{person}{Matthew Goldstein}.} \bibinfo{year}{1979}\natexlab{}.
\newblock \showarticletitle{Discriminant analysis}.
\newblock \bibinfo{journal}{\emph{Biometrics}} \bibinfo{volume}{2},
  \bibinfo{number}{11} (\bibinfo{year}{1979}), \bibinfo{pages}{69--85}.
\newblock


\bibitem[Lazarevic et~al\mbox{.}(2003)]%
        {lazarevic2003}
\bibfield{author}{\bibinfo{person}{Aleksandar Lazarevic},
  \bibinfo{person}{Levent Ertoz}, \bibinfo{person}{Vipin Kumar},
  \bibinfo{person}{Aysel Ozgur}, {and} \bibinfo{person}{Jaideep Srivastava}.}
  \bibinfo{year}{2003}\natexlab{}.
\newblock \showarticletitle{A comparative study of anomaly detection schemes in
  network intrusion detection}. In \bibinfo{booktitle}{\emph{3rd IEEE
  International Conference on Data Mining (ICDM'03)}}.
  \bibinfo{publisher}{IEE}, \bibinfo{pages}{25--36}.
\newblock


\bibitem[Liu et~al\mbox{.}(2009)]%
        {LIU2009}
\bibfield{author}{\bibinfo{person}{Huawen Liu}, \bibinfo{person}{Jigui Sun},
  \bibinfo{person}{Lei Liu}, {and} \bibinfo{person}{Huijie Zhang}.}
  \bibinfo{year}{2009}\natexlab{}.
\newblock \showarticletitle{Feature selection with dynamic mutual information}.
\newblock \bibinfo{journal}{\emph{Pattern Recognition}} \bibinfo{volume}{42},
  \bibinfo{number}{7} (\bibinfo{year}{2009}), \bibinfo{pages}{1330--1339}.
\newblock


\bibitem[Lundberg and Lee(2017)]%
        {Lunberg2017}
\bibfield{author}{\bibinfo{person}{Scott~M. Lundberg} {and}
  \bibinfo{person}{Su-In Lee}.} \bibinfo{year}{2017}\natexlab{}.
\newblock \showarticletitle{A Unified Approach to Interpreting Model
  Predictions (NeurIPS'17)}. In \bibinfo{booktitle}{\emph{Neural Information
  Processing Systems (NeurIPS'17)}}. \bibinfo{pages}{17212--17223}.
\newblock


\bibitem[Moon et~al\mbox{.}(1995)]%
        {Moon1995}
\bibfield{author}{\bibinfo{person}{Young-Il Moon}, \bibinfo{person}{Balaji
  Rajagopalan}, {and} \bibinfo{person}{Upmanu Lall}.}
  \bibinfo{year}{1995}\natexlab{}.
\newblock \showarticletitle{Estimation of mutual information using kernel
  density estimators}.
\newblock \bibinfo{journal}{\emph{Physical Review E}} \bibinfo{volume}{52},
  \bibinfo{number}{3} (\bibinfo{year}{1995}), \bibinfo{pages}{2318--2321}.
\newblock


\bibitem[Moustafa(2021)]%
        {moustafa2021}
\bibfield{author}{\bibinfo{person}{Nour Moustafa}.}
  \bibinfo{year}{2021}\natexlab{}.
\newblock \showarticletitle{A new distributed architecture for evaluating
  AI-based security systems at the edge: Network TON\_IoT datasets}.
\newblock \bibinfo{journal}{\emph{Sustainable Cities and Society}}
  \bibinfo{volume}{72} (\bibinfo{year}{2021}), \bibinfo{pages}{102994}.
\newblock


\bibitem[Moustafa and Slay(2015)]%
        {moustafa17}
\bibfield{author}{\bibinfo{person}{Nour Moustafa} {and} \bibinfo{person}{Jill
  Slay}.} \bibinfo{year}{2015}\natexlab{}.
\newblock \showarticletitle{A hybrid feature selection for network intrusion
  detection systems: Central points}. In \bibinfo{booktitle}{\emph{Australian
  Information Warfare Conference (AIWC'15)}}, Vol.~\bibinfo{volume}{16}.
  \bibinfo{publisher}{Security Research Institute (SRI)},
  \bibinfo{address}{Brisbane}, \bibinfo{pages}{5--13}.
\newblock


\bibitem[Naseer et~al\mbox{.}(2018)]%
        {naseer2018}
\bibfield{author}{\bibinfo{person}{Sheraz Naseer}, \bibinfo{person}{Yasir
  Saleem}, \bibinfo{person}{Shehzad Khalid}, \bibinfo{person}{Muhammad~Khawar
  Bashir}, \bibinfo{person}{Jihun Han}, \bibinfo{person}{Muhammad~Munwar
  Iqbal}, {and} \bibinfo{person}{Kijun Han}.} \bibinfo{year}{2018}\natexlab{}.
\newblock \showarticletitle{Enhanced network anomaly detection based on deep
  neural networks}.
\newblock \bibinfo{journal}{\emph{IEEE Access}}  \bibinfo{volume}{6}
  (\bibinfo{year}{2018}), \bibinfo{pages}{48231--48246}.
\newblock


\bibitem[Neyman and Pearson(1933)]%
        {neyman_pearson_1933}
\bibfield{author}{\bibinfo{person}{Jerzy Neyman} {and} \bibinfo{person}{Egon
  Pearson}.} \bibinfo{year}{1933}\natexlab{}.
\newblock \showarticletitle{The testing of statistical hypotheses in relation
  to probabilities a priori}.
\newblock \bibinfo{journal}{\emph{Mathematical Proceedings of the Cambridge
  Philosophical Society}} \bibinfo{volume}{29}, \bibinfo{number}{4}
  (\bibinfo{year}{1933}), \bibinfo{pages}{492–510}.
\newblock


\bibitem[Paluš(1996)]%
        {Plaus1996}
\bibfield{author}{\bibinfo{person}{Milan Paluš}.}
  \bibinfo{year}{1996}\natexlab{}.
\newblock \showarticletitle{Detecting nonlinearity in multivariate time
  series}.
\newblock \bibinfo{journal}{\emph{Physics Letters A}} \bibinfo{volume}{213},
  \bibinfo{number}{3} (\bibinfo{year}{1996}), \bibinfo{pages}{138--147}.
\newblock


\bibitem[Paninski(2003)]%
        {Paninski2007}
\bibfield{author}{\bibinfo{person}{Liam Paninski}.}
  \bibinfo{year}{2003}\natexlab{}.
\newblock \showarticletitle{Estimation of Entropy and Mutual Information}.
\newblock \bibinfo{journal}{\emph{Neural Computation}} \bibinfo{volume}{15},
  \bibinfo{number}{6} (\bibinfo{year}{2003}), \bibinfo{pages}{1191–1253}.
\newblock


\bibitem[Pearson(1901)]%
        {pearson1901}
\bibfield{author}{\bibinfo{person}{Karl Pearson}.}
  \bibinfo{year}{1901}\natexlab{}.
\newblock \showarticletitle{LIII. On lines and planes of closest fit to systems
  of points in space}.
\newblock \bibinfo{journal}{\emph{The London, Edinburgh, and Dublin
  Philosophical Magazine and Journal of Science}} \bibinfo{volume}{2},
  \bibinfo{number}{11} (\bibinfo{year}{1901}), \bibinfo{pages}{559--572}.
\newblock


\bibitem[Peng et~al\mbox{.}(2005)]%
        {peng05}
\bibfield{author}{\bibinfo{person}{Hanchuan Peng}, \bibinfo{person}{Fuhui
  Long}, {and} \bibinfo{person}{Chris Ding}.} \bibinfo{year}{2005}\natexlab{}.
\newblock \showarticletitle{Feature selection based on mutual information
  criteria of max-dependency, max-relevance, and min-redundancy}.
\newblock \bibinfo{journal}{\emph{IEEE Transactions on Pattern Analysis and
  Machine Intelligence}} \bibinfo{volume}{27}, \bibinfo{number}{8}
  (\bibinfo{year}{2005}), \bibinfo{pages}{1226--1238}.
\newblock


\bibitem[Poole et~al\mbox{.}(2019)]%
        {Poole2019}
\bibfield{author}{\bibinfo{person}{Ben Poole}, \bibinfo{person}{Sherjil Ozair},
  \bibinfo{person}{Aaron Oord}, \bibinfo{person}{Alexander Alemi}, {and}
  \bibinfo{person}{George Tucker}.} \bibinfo{year}{2019}\natexlab{}.
\newblock \showarticletitle{On Variational Bounds of Mutual Information}. In
  \bibinfo{booktitle}{\emph{37th International Conference of Machine Learning
  (ICML'19)}}. \bibinfo{publisher}{PMLR}, \bibinfo{pages}{5171--5180}.
\newblock


\bibitem[Purnama et~al\mbox{.}(2020)]%
        {purnama20}
\bibfield{author}{\bibinfo{person}{Benni Purnama}, \bibinfo{person}{Eko~Arip
  Winanto}, \bibinfo{person}{Deris Stiawan}, \bibinfo{person}{Darmawiiovo
  Hanapi}, \bibinfo{person}{Mohd~Yazid bin Idris}, \bibinfo{person}{Rahmat
  Budiarto}, {et~al\mbox{.}}} \bibinfo{year}{2020}\natexlab{}.
\newblock \showarticletitle{Features extraction on IoT intrusion detection
  system using principal components analysis (PCA)}. In
  \bibinfo{booktitle}{\emph{7th International Conference on Electrical
  Engineering, Computer Sciences and Informatics (EECSI'20)}}.
  \bibinfo{publisher}{IEEE}, \bibinfo{pages}{114--118}.
\newblock


\bibitem[Reboredo et~al\mbox{.}(2016)]%
        {reboredo16}
\bibfield{author}{\bibinfo{person}{Hugo Reboredo}, \bibinfo{person}{Francesco
  Renna}, \bibinfo{person}{Robert Calderbank}, {and} \bibinfo{person}{Miguel~RD
  Rodrigues}.} \bibinfo{year}{2016}\natexlab{}.
\newblock \showarticletitle{Bounds on the number of measurements for reliable
  compressive classification}.
\newblock \bibinfo{journal}{\emph{IEEE Transactions on Signal Processing}}
  \bibinfo{volume}{64}, \bibinfo{number}{22} (\bibinfo{year}{2016}),
  \bibinfo{pages}{5778--5793}.
\newblock


\bibitem[Russell and Norvig(2020)]%
        {russell2010artificial}
\bibfield{author}{\bibinfo{person}{Stuart Russell} {and} \bibinfo{person}{Peter
  Norvig}.} \bibinfo{year}{2020}\natexlab{}.
\newblock \bibinfo{booktitle}{\emph{Artificial Intelligence: A Modern Approach}
  (\bibinfo{edition}{fourth} ed.)}.
\newblock \bibinfo{publisher}{Prentice Hall Press}, \bibinfo{address}{USA}.
\newblock


\bibitem[Schreiber(2000)]%
        {schreiber2000}
\bibfield{author}{\bibinfo{person}{Thomas Schreiber}.}
  \bibinfo{year}{2000}\natexlab{}.
\newblock \showarticletitle{Measuring information transfer}.
\newblock \bibinfo{journal}{\emph{Physical Review Letters}}
  \bibinfo{volume}{85}, \bibinfo{number}{2} (\bibinfo{year}{2000}),
  \bibinfo{pages}{461--464}.
\newblock


\bibitem[Selvakumar and Muneeswaran(2019)]%
        {selvakumar19}
\bibfield{author}{\bibinfo{person}{B Selvakumar} {and}
  \bibinfo{person}{Karuppiah Muneeswaran}.} \bibinfo{year}{2019}\natexlab{}.
\newblock \showarticletitle{Firefly algorithm based feature selection for
  network intrusion detection}.
\newblock \bibinfo{journal}{\emph{Computers \& Security}}  \bibinfo{volume}{81}
  (\bibinfo{year}{2019}), \bibinfo{pages}{148--155}.
\newblock


\bibitem[Shannon(1948)]%
        {shannon1948}
\bibfield{author}{\bibinfo{person}{Claude~E Shannon}.}
  \bibinfo{year}{1948}\natexlab{}.
\newblock \showarticletitle{A mathematical theory of communication}.
\newblock \bibinfo{journal}{\emph{The Bell System Technical Journal}}
  \bibinfo{volume}{27}, \bibinfo{number}{3} (\bibinfo{year}{1948}),
  \bibinfo{pages}{379--423}.
\newblock


\bibitem[Shapley(1953)]%
        {Shapley1953}
\bibfield{author}{\bibinfo{person}{Lloyd Shapley}.}
  \bibinfo{year}{1953}\natexlab{}.
\newblock \showarticletitle{A Value for n-Person Games}.
\newblock \bibinfo{journal}{\emph{Contributions to the Theory of Games}}
  \bibinfo{volume}{2}, \bibinfo{number}{28} (\bibinfo{year}{1953}),
  \bibinfo{pages}{307--318}.
\newblock


\bibitem[Sharafaldin et~al\mbox{.}(2018)]%
        {sharafaldin2018}
\bibfield{author}{\bibinfo{person}{Iman Sharafaldin},
  \bibinfo{person}{Arash~Habibi Lashkari}, {and} \bibinfo{person}{Ali~A
  Ghorbani}.} \bibinfo{year}{2018}\natexlab{}.
\newblock \showarticletitle{Toward generating a new intrusion detection dataset
  and intrusion traffic characterization.}. In \bibinfo{booktitle}{\emph{4th
  International Conference on Information Systems Security and Privacy
  (ICISSP'18)}}. \bibinfo{pages}{108--116}.
\newblock


\bibitem[Sharafaldin et~al\mbox{.}(2019)]%
        {sharafaldin2019}
\bibfield{author}{\bibinfo{person}{Iman Sharafaldin},
  \bibinfo{person}{Arash~Habibi Lashkari}, \bibinfo{person}{Saqib Hakak}, {and}
  \bibinfo{person}{Ali~A Ghorbani}.} \bibinfo{year}{2019}\natexlab{}.
\newblock \showarticletitle{Developing realistic distributed denial of service
  (DDoS) attack dataset and taxonomy}. In \bibinfo{booktitle}{\emph{2019
  International Carnahan Conference on Security Technology (ICCST'19)}}.
  \bibinfo{publisher}{IEEE}, \bibinfo{pages}{1--8}.
\newblock


\bibitem[Shone et~al\mbox{.}(2018)]%
        {shone2018}
\bibfield{author}{\bibinfo{person}{Nathan Shone}, \bibinfo{person}{Tran~Nguyen
  Ngoc}, \bibinfo{person}{Vu~Dinh Phai}, {and} \bibinfo{person}{Qi Shi}.}
  \bibinfo{year}{2018}\natexlab{}.
\newblock \showarticletitle{A deep learning approach to network intrusion
  detection}.
\newblock \bibinfo{journal}{\emph{IEEE Transactions on Emerging Topics in
  Computational Intelligence}} \bibinfo{volume}{2}, \bibinfo{number}{1}
  (\bibinfo{year}{2018}), \bibinfo{pages}{41--50}.
\newblock


\bibitem[Subba et~al\mbox{.}(2016)]%
        {subba2016}
\bibfield{author}{\bibinfo{person}{Basant Subba}, \bibinfo{person}{Santosh
  Biswas}, {and} \bibinfo{person}{Sushanta Karmakar}.}
  \bibinfo{year}{2016}\natexlab{}.
\newblock \showarticletitle{A neural network based system for intrusion
  detection and attack classification}. In \bibinfo{booktitle}{\emph{National
  Conference on Communication (NCC'16)}}. \bibinfo{pages}{1--6}.
\newblock


\bibitem[Sun et~al\mbox{.}(2020)]%
        {sun2020}
\bibfield{author}{\bibinfo{person}{Pengfei Sun}, \bibinfo{person}{Pengju Liu},
  \bibinfo{person}{Qi Li}, \bibinfo{person}{Chenxi Liu},
  \bibinfo{person}{Xiangling Lu}, \bibinfo{person}{Ruochen Hao}, {and}
  \bibinfo{person}{Jinpeng Chen}.} \bibinfo{year}{2020}\natexlab{}.
\newblock \showarticletitle{DL-IDS: Extracting features using CNN-LSTM hybrid
  network for intrusion detection system}.
\newblock \bibinfo{journal}{\emph{Security and Communication Networks}}
  \bibinfo{volume}{20} (\bibinfo{year}{2020}), \bibinfo{pages}{1--11}.
\newblock


\bibitem[Sundararajan and Najmi(2020)]%
        {Sundararajan20}
\bibfield{author}{\bibinfo{person}{Mukund Sundararajan} {and}
  \bibinfo{person}{Amir Najmi}.} \bibinfo{year}{2020}\natexlab{}.
\newblock \showarticletitle{The Many Shapley Values for Model Explanation}. In
  \bibinfo{booktitle}{\emph{37th International Conference on Machine Learning
  (ICML'20)}}. \bibinfo{publisher}{PMLR}, \bibinfo{pages}{9269--9278}.
\newblock


\bibitem[Tavallaee et~al\mbox{.}(2009)]%
        {tavallaee2009}
\bibfield{author}{\bibinfo{person}{Mahbod Tavallaee}, \bibinfo{person}{Ebrahim
  Bagheri}, \bibinfo{person}{Wei Lu}, {and} \bibinfo{person}{Ali~A Ghorbani}.}
  \bibinfo{year}{2009}\natexlab{}.
\newblock \showarticletitle{A detailed analysis of the KDD CUP 99 data set}. In
  \bibinfo{booktitle}{\emph{2009 IEEE Symposium on Computational Intelligence
  for Security and Defense Applications}}. \bibinfo{publisher}{IEEE},
  \bibinfo{pages}{1--6}.
\newblock


\bibitem[Tibshirani(1996)]%
        {tibshirani1996regression}
\bibfield{author}{\bibinfo{person}{Robert Tibshirani}.}
  \bibinfo{year}{1996}\natexlab{}.
\newblock \showarticletitle{{Regression shrinkage and selection via the
  Lasso}}.
\newblock \bibinfo{journal}{\emph{Journal of the Royal Statistical Society
  Series B: Statistical Methodology}} \bibinfo{volume}{58}, \bibinfo{number}{1}
  (\bibinfo{year}{1996}), \bibinfo{pages}{267--288}.
\newblock


\bibitem[van~den Oord et~al\mbox{.}(2019)]%
        {oord2018}
\bibfield{author}{\bibinfo{person}{Aaron van~den Oord}, \bibinfo{person}{Yazhe
  Li}, {and} \bibinfo{person}{Oriol Vinyals}.} \bibinfo{year}{2019}\natexlab{}.
\newblock \bibinfo{title}{Representation Learning with Contrastive Predictive
  Coding}.
\newblock
\newblock
\showeprint[arxiv]{1807.03748}~[cs.LG]


\bibitem[Verma and Ranga(2019)]%
        {verma2019}
\bibfield{author}{\bibinfo{person}{Abhishek Verma} {and}
  \bibinfo{person}{Virender Ranga}.} \bibinfo{year}{2019}\natexlab{}.
\newblock \showarticletitle{ELNIDS: Ensemble learning based network intrusion
  detection system for RPL based Internet of Things}. In
  \bibinfo{booktitle}{\emph{4th International Conference on Internet of Things:
  Smart Innovation and Usages (IoT-SIU'19)}}. \bibinfo{publisher}{IEEE}.
\newblock


\bibitem[Williams and Beer(2010)]%
        {williams2010}
\bibfield{author}{\bibinfo{person}{Paul~L. Williams} {and}
  \bibinfo{person}{Randall~D. Beer}.} \bibinfo{year}{2010}\natexlab{}.
\newblock \bibinfo{title}{Nonnegative Decomposition of Multivariate
  Information}.
\newblock
\newblock
\showeprint[arxiv]{1004.2515}~[cs.IT]


\bibitem[Wollstadt et~al\mbox{.}(2023)]%
        {wollstadt23}
\bibfield{author}{\bibinfo{person}{Patricia Wollstadt},
  \bibinfo{person}{Sebastian Schmitt}, {and} \bibinfo{person}{Michael Wibral}.}
  \bibinfo{year}{2023}\natexlab{}.
\newblock \showarticletitle{A rigorous information-theoretic definition of
  redundancy and relevancy in feature selection based on (partial) information
  decomposition}.
\newblock \bibinfo{journal}{\emph{Journal of Machine Learning Research}}
  \bibinfo{volume}{24}, \bibinfo{number}{131} (\bibinfo{year}{2023}),
  \bibinfo{pages}{1--44}.
\newblock


\bibitem[Xu et~al\mbox{.}(2018)]%
        {xu2018}
\bibfield{author}{\bibinfo{person}{Congyuan Xu}, \bibinfo{person}{Jizhong
  Shen}, \bibinfo{person}{Xin Du}, {and} \bibinfo{person}{Fan Zhang}.}
  \bibinfo{year}{2018}\natexlab{}.
\newblock \showarticletitle{An intrusion detection system using a deep neural
  network with gated recurrent units}.
\newblock \bibinfo{journal}{\emph{IEEE Access}}  \bibinfo{volume}{6}
  (\bibinfo{year}{2018}), \bibinfo{pages}{48697--48707}.
\newblock


\bibitem[Yamada et~al\mbox{.}(2020)]%
        {yamada20}
\bibfield{author}{\bibinfo{person}{Yutaro Yamada}, \bibinfo{person}{Ofir
  Lindenbaum}, \bibinfo{person}{Sahand Negahban}, {and} \bibinfo{person}{Yuval
  Kluger}.} \bibinfo{year}{2020}\natexlab{}.
\newblock \showarticletitle{Feature Selection using Stochastic Gates}. In
  \bibinfo{booktitle}{\emph{37th International Conference on Machine Learning
  (ICML'20)}}. \bibinfo{publisher}{PMLR}, \bibinfo{pages}{10648--10659}.
\newblock


\bibitem[Yin et~al\mbox{.}(2017)]%
        {yin2017}
\bibfield{author}{\bibinfo{person}{Chuanlong Yin}, \bibinfo{person}{Yuefei
  Zhu}, \bibinfo{person}{Jinlong Fei}, {and} \bibinfo{person}{Xinzheng He}.}
  \bibinfo{year}{2017}\natexlab{}.
\newblock \showarticletitle{A deep learning approach for intrusion detection
  using recurrent neural networks}.
\newblock \bibinfo{journal}{\emph{IEEE Access}}  \bibinfo{volume}{5}
  (\bibinfo{year}{2017}), \bibinfo{pages}{21954--21961}.
\newblock


\bibitem[Zeng et~al\mbox{.}(2019)]%
        {zeng2019}
\bibfield{author}{\bibinfo{person}{Yi Zeng}, \bibinfo{person}{Huaxi Gu},
  \bibinfo{person}{Wenting Wei}, {and} \bibinfo{person}{Yantao Guo}.}
  \bibinfo{year}{2019}\natexlab{}.
\newblock \showarticletitle{$ Deep-Full-Range $: a deep learning based network
  encrypted traffic classification and intrusion detection framework}.
\newblock \bibinfo{journal}{\emph{IEEE Access}}  \bibinfo{volume}{7}
  (\bibinfo{year}{2019}), \bibinfo{pages}{45182--45190}.
\newblock


\end{thebibliography}


\newpage

\appendix

\section{FSNID Algorithm}\label{app:fsnid_alg}
In this section, we present Algorithm \ref{alg:algorithm}, which formally defines FSNID.
\begin{algorithm}[!t]
    \caption{FSNID.}
    \label{alg:algorithm}
    \textbf{Input}: Network features $\mathcal{X}$ and ground truth labels $Y$, respectively (see Section \ref{sec:preliminaries}). \textbf{Output}: ${\mathcal{X}}_*$ (a desirable subset of features).
    \begin{algorithmic}[1]
    \vspace{-10pt}
    \STATE Initialize ${\mathcal{X}}_* = \{\}$
    \FOR{$i = 1$ to $N$}
        \IF{$\Phi_{{X}_i; {\mathcal{X}} \rightarrow {Y}} = 0$} 
                \STATE ${\mathcal{X}} = {\mathcal{X}} \backslash \{{X}_i\}$
        \ELSE
            \STATE ${\mathcal{X}}_* = {\mathcal{X}}_* \cup \{{X}_i\}$
        \ENDIF
    \ENDFOR
    \STATE \textbf{return} ${\mathcal{X}}_*$
    \end{algorithmic}
\end{algorithm}

\section{Performance Guarantees}\label{app:conv_gar}
In this section, we intend to demonstrate that a classifier trained on the set of features selected by FSNID will converge to an optimal solution. To begin, we first note that according to \cite{reboredo16} the MI between the target and the feature set bounds the performance of a classifier. Consequently, we aim to show that $I(Y;\mathcal{X})=I(Y;\mathcal{X}_*)$, and therefore, the bounds on performance remain the same whether using the full or selected set of features. 

To begin, we present a Lemma that demonstrates the three types of variables that satisfy $\Phi_{{X}_i;{\mathcal{X}}\rightarrow {Y}}=0$.

\noindent  \textbf{Lemma 1.} \textit{Irrelevant, redundant, and perfectly redundant variables satisfy $\Phi_{{X}_i;{\mathcal{X}}\rightarrow {Y}}=0$. More specifically:
\begin{equation}
\begin{split}
  & \Phi_{{X}_i;{\mathcal{X}}\rightarrow {Y}} =0\Leftrightarrow \\ & (X_i \in \mathcal{X}, \forall \mathcal{P} \in \mathscr{P}(\mathcal{X}):H(Y|\mathcal{P})=H(Y|\mathcal{P}\cup X_i)) \\ & \text{{(Irrelevant)}} \\ & \text{or} \quad (X_i \in \mathcal{X}:H(Y|\mathcal{X})=H(Y|\mathcal{X}_{\backslash X_i})<H(Y|X_i)<H(Y)) \\ & \text{{(Redundant)}} \\ & 
  \text{or} \quad (X_i \in \mathcal{X}:H(Y|\mathcal{X})=H(Y|\mathcal{X}_{\backslash X_i})=H(Y|X_i)<H(Y)) \\ & \text{{(Perfectly Redundant)}}
\end{split}
\label{eqn:lemma1}
\end{equation}}
\noindent   \textit{Proof.} To prove this, we first note that $\Phi_{{X}_i;{\mathcal{X}}\rightarrow {Y}}=0 \Leftrightarrow H(Y|\mathcal{X})=H(Y|\mathcal{X}_{\backslash X_i})$. However, the equality $H(Y|\mathcal{X})=H(Y|\mathcal{X}_{\backslash X_i})$ reveals no information regarding the following relationship: $H(Y|X_i) \leq H(Y)$. This gives us two options $H(Y|X_i) = H(Y)$ or $H(Y|X_i) < H(Y)$. We first present the latter, which corresponds to the irrelevant variable case. As stated, let us assume $H(Y|X_i) = H(Y)$; therefore, $H(Y|X_i) = H(Y) \quad \& \quad H(Y|\mathcal{X})=H(Y|\mathcal{X}_{\backslash X_i})$. This leads us too:
\begin{equation}
\begin{split}
 (H(Y|&X_i) = H(Y)  \quad \& \quad H(Y|\mathcal{X})=H(Y|\mathcal{X}_{\backslash X_i})) \\ &
\Leftrightarrow (X_i \in \mathcal{X}, \forall \mathcal{P} \in \ \{ \varnothing, \mathcal{X}_{\backslash X_i} \} :H(Y|\mathcal{P})=H(Y|\mathcal{P}\cup X_i))  \\ &
 \Leftrightarrow (X_i \in \mathcal{X}, \forall \mathcal{P} \in \mathscr{P}(\mathcal{X}):H(Y|\mathcal{P})=H(Y|\mathcal{P}\cup X_i)).
\end{split}
\label{eqn:lemma1_2}
\end{equation}
\noindent therefore proving Lemma 1 for irrelevant variables. We now provide a similar analysis for the redundant and perfectly redundant case. Let us now suppose that $H(Y|X_i) < H(Y)$, this leads to:
\begin{equation}
\begin{split}
 (H(Y|&X_i) < H(Y) \quad \& \quad ( X_i \in \mathcal{X}: H(Y|\mathcal{X})=H(Y|\mathcal{X}_{\backslash X_i})) \\ &
 \Leftrightarrow ( X_i \in \mathcal{X}: H(Y|\mathcal{X})=H(Y|\mathcal{X}_{\backslash X_i})\leq H(Y|X_i)<H(Y)) \\ &
 \Leftrightarrow ( X_i \in \mathcal{X}: H(Y|\mathcal{X})=H(Y|\mathcal{X}_{\backslash X_i})< H(Y|X_i)<H(Y) \\ &
 \text{or} \quad H(Y|\mathcal{X})=H(Y|\mathcal{X}_{\backslash X_i}) = H(Y|X_i)<H(Y)).
\end{split}
\label{eqn:lemma1_3}
\end{equation}
Therefore, we have proved Lemma 1 in the redundant or perfectly redundant variable case.

Via Lemma 1 we demonstrated that if $\Phi_{{X}_i;{\mathcal{X}}\rightarrow {Y}}=0$ is true if features are \textit{irrelevant, redundant} or \textit{perfectly redundant}. We now prove that, by removing \textit{irrelevant} and \textit{redundant} features we lose no information regarding the target:

\noindent    \textbf{Lemma 2.} \textit{Removing irrelevant or redundant variables (as defined in Equation \ref{eqn:lemma1}) from the set of features does not diminish the resulting sets ability to reduce the uncertainty of the target. More formally, for the case of irrelevant variables the following holds:
\begin{equation}
\begin{split}
  & H(Y|\{X_i \in \mathcal{X} \exists \mathcal{P} \in \mathscr{P}(\mathcal{X}): H(Y|\mathcal{P})>H(Y|\mathcal{P}\cup X_i) \}) = H(Y|\mathcal{X}),
\end{split}
\label{eqn:lemma2_1}
\end{equation}
and for the case of redundant variables, the following holds:
\begin{equation}
\begin{split}
  & H(Y|\{X_i \in \mathcal{X}: \neg(H(Y|\mathcal{X})=H(Y|\mathcal{X}_{\backslash X_i})<H(Y|X_i)<H(Y)) \}) \\ & = H(Y|\mathcal{X})
\end{split}
\label{eqn:lemma2_2}
\end{equation}
}
\noindent    \textit{Proof.} We begin by proving the irrelevant case. The set $\mathcal{X}$ can be divided into irrelevant $\mathcal{X}_{I} = \{X_{1,I},X_{2,I} \dots X_{N^{I},I} \}$ and non-irrelevant features $\mathcal{X}_{NI} = \{X_{1,NI},X_{2,NI} \dots X_{N^{NI},NI} \}$. Given, $\mathcal{X}_{NI} = \{X_i \in \mathcal{X} \exists \mathcal{P} \in \mathscr{P}(\mathcal{X}): H(Y|\mathcal{P})>H(Y|\mathcal{P}\cup X_i) \}$ and $\mathcal{X}=\mathcal{X}_{NI} \cup \mathcal{X}_{I}$ we now demonstrate that $H(Y|\mathcal{X}_{NI})=H(Y|\mathcal{X})$. To begin, we write out the condition to which irrelevant variables are subjected. If $X_i \in \mathcal{X}_I$ we can write:
\begin{equation}
\begin{split}
  & X_i \in \mathcal{X}, \forall \mathcal{P} \in \mathscr{P}(\mathcal{X}):H(Y|\mathcal{P})=H(Y|\mathcal{P}\cup X_i) \\ & \text{{(By substituting $\mathscr{P}(\mathcal{X}) = \mathcal{X}_{NI}$ we get:) }} \\ & X_i \in \mathcal{X}(\mathcal{X}):H(Y|\mathcal{X}_{NI})=H(Y|\mathcal{X}_{NI}\cup X_i). 
\end{split}
\label{eqn:lemma2_3}
\end{equation}
However, it is also true that:
\begin{equation}
\begin{split}
  & \text{{(By substituting $\mathscr{P}(\mathcal{X}) = \mathcal{X}_{NI} \cup X_i $ we get:)}} \\ & X_i \in \mathcal{X}(\mathcal{X}):H(Y|\mathcal{X}_{NI}\cup X_i)=H(Y|\mathcal{X}_{NI}\cup X_i \cup X_i).  
\end{split}
\label{eqn:lemma2_32}
\end{equation}
In Equations \ref{eqn:lemma2_3} and \ref{eqn:lemma2_32}, we have shown that $H(Y|\mathcal{X}_{NI}) = H(Y|\mathcal{X}_{NI}\cup X_i \cup X_i)$. However, the process above can be applied indefinitely for all $X_i \in \mathcal{X}_I$ to show: $H(Y|\mathcal{X}_{NI})=H(Y|\mathcal{X})$. We now move on to the redundant variables case. As described earlier, redundant features are those that satisfy: $H(Y|\mathcal{X})=H(Y|\mathcal{X}_{\backslash X_i})<H(Y|X_i)<H(Y)) $. To complete this proof, we separate our set $\mathcal{X}$ into its redundant  $\mathcal{X}_{R} = \{X_{1,R},X_{2,R} \dots X_{N^{R},R} \}$ and non redundant parts $\mathcal{X}_{NR} = \{X_{1,NR},  X_{2,NR} \dots X_{N^{NR},NR} \}$, where $\mathcal{X}_{NR} = \{X_i \in \mathcal{X}: \neg(H(Y|\mathcal{X})=H(Y|\mathcal{X}_{\backslash X_i})<H(Y|X_i)<H(Y)) \}$ and $\mathcal{X}=\mathcal{X}_{NR} \cup \mathcal{X}_{R}$. Given our definition of redundant features, the following holds:
\begin{equation}
\begin{split}
     & H(Y|\mathcal{X}) = 
    H(Y|\mathcal{X}_{\backslash X_{1,R}}) = H(Y|\mathcal{X}_{\backslash X_{2,R}}) = \dots H(Y|\mathcal{X}_{\backslash X_{N^R,R}}) \\ &
  \text{(By monotonicity of conditional entropy we get:)} \\ & = H(Y|\mathcal{X}_{\backslash X_{1,R}} \cap 
 \mathcal{X}_{\backslash X_{2,R}} \dots \cap \mathcal{X}_{\backslash X_{N^R,R}}) \\ & = H(Y|\mathcal{X}_{NR}).
\end{split}
\label{eqn:lemma2_4}
\end{equation}

\noindent \textbf{Lemma 3.} \textit{Removing perfectly redundant variables from the set of features diminishes the extent by which the resulting set reduces the uncertainty of the target. More formally:
\begin{equation}
\begin{split}
  & H(Y|\{X_i \in \mathcal{X}: \neg(H(Y|\mathcal{X})=H(Y|\mathcal{X}_{\backslash X_i})=H(Y|X_i)<H(Y)) \}) \\ & \neq H(Y|\mathcal{X})
\end{split}
\label{eqn:lemma3}
\end{equation}
} \noindent  \textit{Proof.} We prove this by showing that if we try and apply the logic followed in Equation \ref{eqn:lemma2_4} from redundant variables to perfectly redundant variables a contradiction arises. In accordance with our definition of perfectly redundant features, the following holds:
\begin{equation}
\begin{split}
     & H(Y|\mathcal{X}) = 
    H(Y|\mathcal{X}_{\backslash X_i}) = H(Y|X_i)  \\ &
  \text{(Attempting to apply monotonicity of conditional entropy we get:)} \\ & = H(Y|\mathcal{X}_{\backslash X_i} \cap 
 X_i) \\ & = H(Y|\varnothing) \\ & = H(Y).
\end{split}
\label{eqn:lemma3_1}
\end{equation}
However, our definition of a perfectly redundant features satisfy: $H(Y|\mathcal{X})=H(Y|\mathcal{X}_{\backslash X_i})=H(Y|X_i)<H(Y)$. Hence, a contradiction has arisen. Consequently, we cannot remove all perfectly redundant features from our set and maintain the extent to which the resulting set reduces the uncertainty of the target.

In Equation \ref{eqn:lemma3_1}, the application of the monotonicity of conditional entropy is no longer valid; however, this is not because it is no longer true. Rather, it is because motonicity applies to ordered sets \cite{russell2010artificial}. In the above, we have the sets $\mathcal{X}_{\backslash X_i}$ and $\{X_i\}$; the notion of ordering these two sets, with no overlap in contents, is ill defined. Consequently, the monotonicity of the function is also ill defined, leading to the result in Equation \ref{eqn:lemma3_1}.

Thus far, we have shown that variables that satisfy $\Phi_{{X}_i; {\mathcal{X}} \rightarrow {Y}} = 0$, will be irrelevant, redundant or perfectly redundant (Lemma 1). Of these, only perfectly redundant features cannot be removed without affecting the extent by which the set of selected features reduces the uncertainty of target variable (Lemmas 2 and 3). We now prove that the simple sequential application in Algorithm \ref{alg:algorithm} overcomes cases of perfect redundancy, so that the resulting set reduces the uncertainty of the target identically to the full set. To do this, we work through an example of our algorithm encountering and removing perfectly redundant variables.

\begin{algorithm*}[t!]
    \caption{TE measure estimation.}
    \label{alg:te_est}
    \textbf{Input}: Training dataset $  ({i}^1, {y}^1, {i}^2, {y}^2... {i}^T, {y}^T) $, and feature of interest ${X}_i$.\\
    \textbf{Output}:  $H({Y}|{\mathcal{X}}_{\backslash {X}_i}) - H({Y}|{\mathcal{X}})$
    \begin{algorithmic}[1] 
        \STATE Initialize weights for $\theta$ and $\theta_{\backslash {X}_i}$
        \FOR{$1$ to $N$}
        \STATE Draw mini batch samples of length $b$ from the joint distribution of the target and the network features with all possible variables included $ p_{Y,\mathcal{X}} \sim (y^{t_1},x^{t_1}_1,x^{t_1}_i \dots  x^{t_1}_N),\dots,(y^{t_b},x^{t_b}_1,x^{t_b}_i \dots  x^{t_b}_N)$, and repeat for the marginal distribution $p_{Y}\otimes p_{\mathcal{X}} \sim (y^{t'_1},x^{t_1}_1,x^{t_1}_i \dots  x^{t_1}_N),\dots,(y^{t'_b},x^{t_b}_1,x^{t_b}_i \dots  x^{t_b}_N)$, where $t'_i \neq t_i$.
        \STATE Draw mini batch samples of length $b$ from the joint distribution of the target and the network features with variable $X_i$ missing.  $ p_{Y,\mathcal{X}_{\backslash X_i}} \sim (y^{t_1},x^{t_1}_1 \dots  x^{t_1}_N),\dots,(y^{t_b},x^{t_b}_1 \dots  x^{t_b}_N)$, and repeat for the marginal distribution $p_{Y}\otimes p_{\mathcal{X}_{\backslash X_i}} \sim (y^{t'_1},x^{t_1}_1, \dots  x^{t_1}_N),\dots,(y^{t'_b},x^{t_b}_1 \dots  x^{t_b}_N)$, where $t'_i \neq t_i$.
        \STATE $I({Y};{\mathcal{X}}) \geq \frac{1}{b}\sum^b_{j=1}F_{\theta}(y^{t_j},x^{t_j}_1,x^{t_j}_i \dots  x^{t_j}_N)) - \frac{1}{b}\sum^b_{j=1}\log{e^{F_{\theta}((y^{t_j'}_1,x^{t_j}_1,x^{t_j}_i \dots  x^{t_j}_N)}}$
        \STATE $I({Y};{\mathcal{X}}_{\backslash {X}_i}) \geq \frac{1}{b}\sum^b_{j=1}F_{\theta_{\backslash {X}_i}}((y^{t_j},x^{t_j}_1 \dots  x^{t_j}_N)) - \frac{1}{b}\sum^b_{j=1}\log{e^{F_{\theta_{\backslash {X}_i}}(y^{t_j'}_1,x^{t_j}_1 \dots  x^{t_j}_N)}}$
        \STATE $\theta \leftarrow \tilde{\nabla}_{\theta} I({Y};{\mathcal{X}})$
        \STATE $\theta_{\backslash {X}_i} \leftarrow \tilde{\nabla}_{\theta_{\backslash {X}_i}} I({Y};{\mathcal{X}}_{\backslash {X}_i})$
        \ENDFOR
        \STATE \textbf{return} $I({Y};{\mathcal{X}}) - I({Y};{\mathcal{X}}_{\backslash {X}_i})$
    \end{algorithmic}
    \caption{Estimation of $\Phi_{{X}_i;{\mathcal{X}} \rightarrow {Y}}$. }
    \label{alg:estimation}
\end{algorithm*}

\noindent \textbf{Theorem 1.} \textit{The set of features selected using Algorithm \ref{alg:algorithm} reduce the uncertainty of the target identically to the full set: $H(Y|\mathcal{X})=H(Y|\mathcal{X}_*)$.}

\noindent \textit{Proof.}
We consider a case of perfect redundancy such that: $H(Y|\mathcal{X})=H(Y|\mathcal{X}_{\backslash X_i})=H(Y|X_i)<H(Y)$. We can write the set ${\mathcal{X}}$ as:  ${\mathcal{X}} = \{\mathcal{X}_{\backslash X_i}, X_i\}$. We can now apply Algorithm \ref{alg:algorithm}: we begin by calculating $\Phi_{X_j; {\mathcal{X}} \rightarrow {A}}$, where $X_j \in \mathcal{X}_{\backslash X_i}$. 
\begin{equation}
\begin{split}
     H(Y|\mathcal{X}_{\backslash X_j}) & =  H(Y|\mathcal{X}_{\backslash \{X_i, X_j\}} \cup \{X_i\}) \\ & \leq H(Y|X_i)  \\ & \quad \text{(by monotonicity of conditional entropy)} \\ & \leq H(Y|\mathcal{X})  \\ & \quad \text{(by definition of perfect redundancy)} \\ H(Y|\mathcal{X}_{\backslash X_j}) - H(Y|\mathcal{X}) &\leq 0 \quad \text{(subtracting $H(Y|\mathcal{X})$)} \\ \Phi_{X_j; {\mathcal{X}} \rightarrow {A}} &\leq 0 \quad \text{(From Equation \ref{eqn:TE_measure})} \\ \Phi_{X_j; {\mathcal{X}} \rightarrow {A}} &= 0 \quad \text{(by non-negativity of our measure\footnote{For proof of non-negativity please refer to Appendix \ref{app:noneg}})}
\end{split}
\label{eqn:theorem_1}
\end{equation}
Consequently, $X_j$ gets removed from the set ${\mathcal{X}}$. However, we have written the above so it applies generally to all $X_j$. Consequently, all $X_j \in \mathcal{X}_{\backslash X_i}$ will be removed from the set for similar reasons. This leaves only $X_i$ in our set $\mathcal{X}$, which according to our definition satisfies $H(Y|\mathcal{X})=H(Y|X_i)$. Therefore, $H(Y|\mathcal{X})=H(Y|\mathcal{X}_*)$ and we no longer lose information due to perfect redundancy. 

Given in Theorem 1, we have demonstrated that $H(Y|\mathcal{X})=H(Y|\mathcal{X}_*)$, it is also true that $I(Y;\mathcal{X}) = I(Y;\mathcal{X}_*)$. Consequently, we can apply the theory presented in \cite{reboredo16}, to ensure that classification using the set $\mathcal{X}_*$ is theoretically as accurate as that achieved using $\mathcal{X}$.

\section{Proof of Non-Negativity of Transfer Entropy Measure}
\label{app:noneg}
In this section, we prove the non-negativity of $\Phi_{{X}_i;{\mathcal{X}} \rightarrow {A}}$. We first write the full expression of the measure as follows: 
\begin{equation}
\label{eqn:noneg1}
\begin{split}
 \Phi_{{X}_i;{\mathcal{X}} \rightarrow {A}} = &  - \int_{{A} \times {\mathcal{X}}} p_{A,\mathcal{X}}(a^t,x^t_1,x^t_i \dots  x^t_N) \\ & \log{\frac{p_{A,\mathcal{X}_{\backslash {X}_i}}(a^t,x^t_1 \dots  x^t_N), p_{\mathcal{X}}( x^t_1,x^t_i \dots  x^t_N)}{p_{A,\mathcal{X}}(a^t,x^t_1,x^t_i \dots  x^t_N), p_{\mathcal{X}_{\backslash {X}_i}}( x^t_1\dots  x^t_N)}} d{A} \times {\mathcal{X}}.
\end{split}
\end{equation}
By applying Jensen's inequality we can then write:
\begin{equation}
\label{eqn:noneg2}
\begin{split}
\Phi_{{X}_i;{\mathcal{X}} \rightarrow {A}}  \geq &  - \log \int_{{A} \times {\mathcal{X}}} p_{A,\mathcal{X}}(a^t,x^t_1,x^t_i, \dots,  x^t_N)\\
& \cdot \frac{p_{A,\mathcal{X}_{\backslash {X}_i}}(a^t,x^t_1, \dots,  x^t_N) \cdot p_{\mathcal{X}}( x^t_1,x^t_i, \dots,  x^t_N)}{p_{A,\mathcal{X}}(a^t,x^t_1,x^t_i, \dots,  x^t_N) \cdot p_{\mathcal{X}_{\backslash {X}_i}}( x^t_1, \dots,  x^t_N)} d{A} \times {\mathcal{X}}\\ 
\geq & - \log \int_{{A} \times {\mathcal{X}}}  \frac{p_{A,\mathcal{X}_{\backslash {X}_i}}(a^t,x^t_1, \dots,  x^t_N)}{p_{\mathcal{X}_{\backslash {X}_i}}( x^t_1, \dots,  x^t_N)} d{A} \times {\mathcal{X}}\\  
\geq & - \log(1) \\  
\geq & 0,
\end{split}
\end{equation}

\noindent therefore proving the non-negativity of $\Phi_{{X}_i;{\mathcal{X}} \rightarrow {A}}$. 

\section{ Derivation of the Transfer Entropy Measure}\label{appendix:derivation}
In this section, we prove the results in Equation \ref{eqn:TE_from_MI}.
\begin{equation}
\begin{split}
   I(Y;\mathcal{X}) - I(Y;\mathcal{X}_{\backslash X_i})  & = H(Y) - H(Y|\mathcal{X}) - (H(Y) - H(Y|\mathcal{X}_{\backslash X_i}) ) \\ & = H(Y|\mathcal{X}_{\backslash X_i}) - H(Y|\mathcal{X}) \\ & = \Phi_{\mathcal{X};X_i \rightarrow Y}
\end{split}
\label{eqn:rel_holds}
\end{equation}

\begin{figure*}[t!]%
    \subfigure{\includegraphics[width=17.5cm]{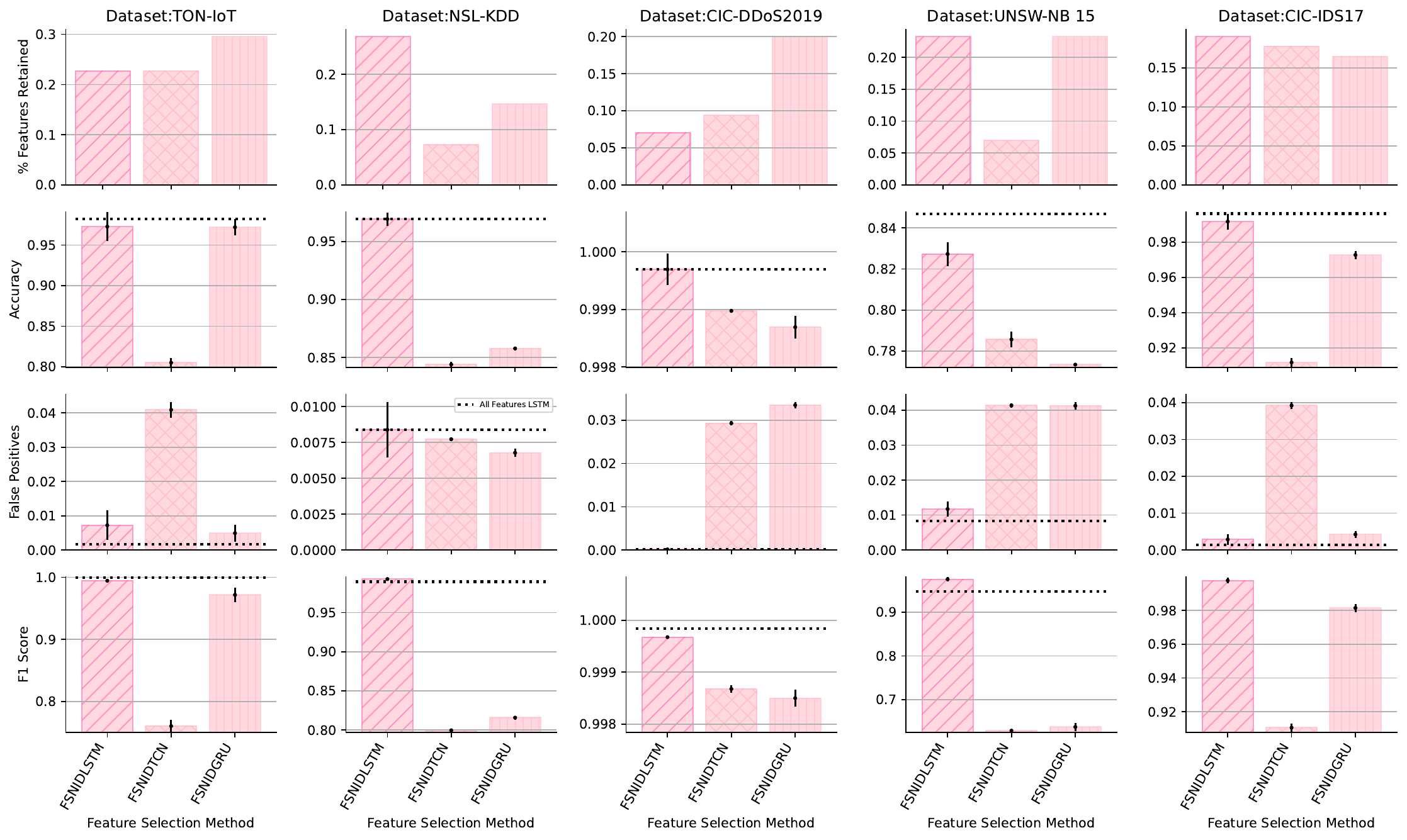} }
    \caption{In this figure we compare three neural architectures ability to incorporate temporal dependencies into the feature selection and subsequent classification tasks. }
    \label{fig:temp_analysis}
\end{figure*}
\section{Recurrent Layer Ablation Studies}\label{app:lstm_ablation}
In this section, we compare the abilities of Temporal Convolutional Networks (TCN), Gated Recurrent Units (GRU), and Long Short-Term Memory networks (LSTM) in capturing temporal dependencies during the NID feature selection and classification tasks. As shown in Figure \ref{fig:temp_analysis}, the LSTM model is the most performant, despite its increased complexity. This complexity results in a slight increase in runtime, as detailed in Appendix \ref{app:runtimes}. However, this extended runtime is not significant when considering the superior performance indicated by the NID results.

\section{Handling the Length-$k$ Problem for Baselines}\label{app:len_k}
In this section, we explain how we systematically place an upper bound on the number of features selected using each baseline, thus avoiding the length-$k$ problem. This step is unnecessary for the LASSO algorithm since it is inherently able to deal with it. For the FSNID method, the procedure is detailed in Section \ref{subsection:practical} and will not be repeated here.
\paragraph{PI}
For the permutation importance, we again utilize a null model. After five experiments, a feature is not selected if its permutation importance does not have a 95\% probability of being higher than that of a random feature.

\paragraph{UMFI}
We adopt a null model as in the PI case. 

\paragraph{CLM} This technique selects features that increase the MI between the target and feature set. Feature addition typically stops when the MI stops rising. To identify this point, we use a null model, adding features as long as the chosen feature increases the MI more than a random feature would.

\paragraph{MIFA} This soft computing technique involves using a metaheuristic to select features that improve MI with the target. Without restricting the number of selected features, the method tends to select all possible features, an undesirable result. To address this, we increase $k$ (the number of possible features) starting from 5 in increments of 5. We then select the smallest feature set for which the next set of features up did not have a statistically significant improvement in MI. This process underscores the challenges of methods that cannot solve the length-$k$ problem, repeating the experiments for multiple values of $k$ exponentially increases temporal complexity. 

\section{In-Depth Dataset Description.}\label{app:dataset}

\noindent \textbf{Bot-IoT.} Specifically curated for IoT networks, the Bot-IoT dataset \cite{koroniotis2019} offers insights into both regular and Botnet cyberattack traffic. Its real-world testbed ensures the data's reliability. This dataset encompasses simulations of five IoT devices, each replicating a different real-world application with 98.8\% attack samples and 1.2\% benign samples.

\noindent \textbf{TON-IoT.} This dataset provides a comprehensive view of network traffic from diverse IoT and IIoT sensors, combined with system traces from both Linux and Windows hosts, improving the generality and diversity of the datastreams available from the Bot-IoT dataset \cite{moustafa2021}. For this study, the dataset from a Windows 10 network was employed, capturing various system activities across 124 attributes. The attack vector in this dataset is comprised of seven unique attack types, accounting for 53.0\% of the target data,
while the remaining 47.0\% is benign.

\noindent \textbf{NSL-KDD.} Designed to address the shortcomings of the KDD'99 dataset \cite{tavallaee2009}, NSL-KDD encompasses 41 features, split into 34 numeric, 4 binary, and 3 nominal attributes. Categorized based on packet, content, traffic, and host characteristics, it classifies attacks into four primary categories: DoS, R2L, U2R, and Probe. These main categories are further subdivided into forty specific attack classes \cite{selvakumar19}, which in total characterize 75.6\% of the total traffic, while the remainder is benign. Unlike both the Bot-IoT and TON-IoT datasets, NSL-KDD contains a large variety of attack styles that are applicable to non-IoT computer networks. 

\noindent \textbf{CIC-DDoS2019.} This dataset focuses on a variety of DDoS attacks, mirroring genuine real-world traffic conditions. Notably, it was generated using data describing the behavior of 25 users across a variety of protocols. For the purpose of our research, we focus on the UDPLag dataset, as this attack type is uncommon, and unique to this dataset \cite{sharafaldin2019}. For this dataset, 98.9\% of the data was benign while the rest were attacks.

\noindent \textbf{UNSW-NB 15.} The UNSW-NB 15 dataset consists of 49 features and nine attack groups, which are associated with 68.1\% of the traffic, while 31.9\% is benign. 
The classification task is particularly challenging for this dataset as it requires the consideration of time dependencies in order to achieve high classification accuracies. 

\noindent \textbf{CICIDS17.} This dataset offers a snapshot of benign traffic with a sparse presence of state-of-the-art attacks, ensuring the data mirrors real-world conditions \cite{sharafaldin2018}. 88 network flow features were extracted from the activities of 25 users considering different protocols. Attacks such as Brute Force FTP and DDoS were incorporated during different times of the day and week, with 63.6\% benign and 36.4\% malicious behaviors. 
\section{Selection of the Hyperparameters}\label{appendix:hyper}
\subsection{Feature Selection Experiments}
When deploying Algorithm \ref{alg:te_est} we used $b = 100$ and $N=10000$ for the vanilla version of FSNID, MIFA, and CLM; whereas, $b = 100$ and $N=20,000$ and sequence length $s$ (the range over which we incorporate temporal dependencies) was set to 10. The differences are due to the more complex temporal relationships requiring extended training to be fully captured. In the vanilla implementation of our method, the function approximator is a standard feedforward multi-layer perceptron, featuring one hidden layer consisting of 50 nodes. For the LSTM-based approach, the architecture remained the same, but it included an additional LSTM layer where the hidden layer also included 50 nodes. This was repeated for GRU and TCN modules. Furthermore, we use a learning rate of $0.0001$ for the calculation of $\Phi_{{X}_i;{\mathcal{X}}\rightarrow {Y}}$ and a learning rate of $0.01$ for the classification task. The networks used for both tasks in both cases where identical, except for MINE (MI neural estimation), we are undertaking a regression task. Therefore, we use a mean squared error loss and an Adam optimizer. Meanwhile, for the classification task we use a Negative Log Likelihood Loss and a standard stochastic gradient descent optimizer. For both PI and UMFI we use 100 random trees and for PI we ran 10 repeats. For LASSO we employed a regularization strength of $\alpha = 0.001$, in order to select enough features. Finally, for the MIFA algorithm we set the number of fireflies to $0.5*N$, where $N$ is the max number of features, and allowed $10$ iterations of brightness reconfiguration.

\subsection{Complexity Analysis Experiments}
For these experiments we use the same hyperparameters as for the feature selection experiments, except in this case we use $b = 10$ and $N=100$ (as defined in Algorithm \ref{alg:algorithm}) for both our vanilla and LSTM-based methods. While the number of random trees in use for our baselines is $10$. These values were chosen, not due to their feature selecting abilities but due to their speed.

\begin{figure*}[t!]%
    \subfigure{\includegraphics[width=17.5cm]{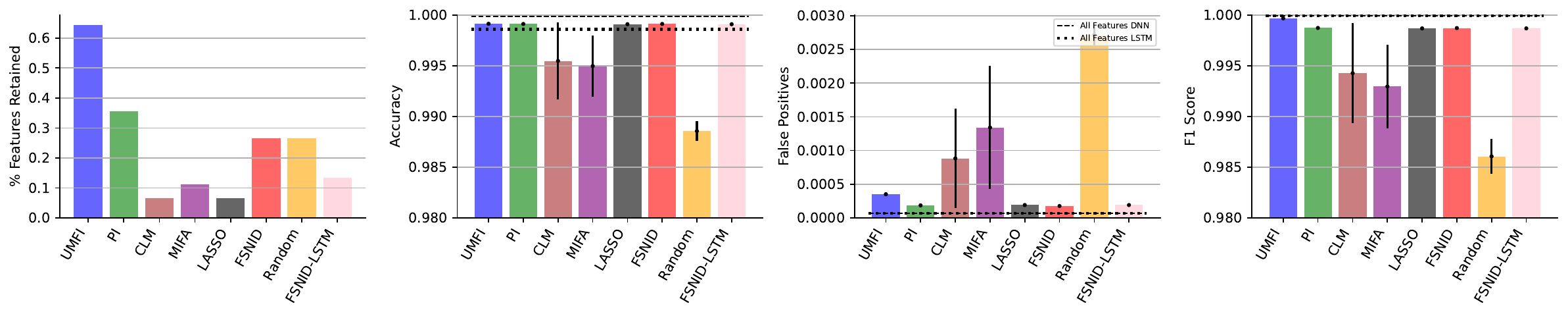} }
    \caption{In this figure we report the results of FSNID and baselines when applied to the Bot-IoT dataset. }
    \label{fig:Botiot_analysis}
\end{figure*}
\section{Further Experiments on Bot-IoT Dataset}\label{app:Botiot_expts}
In this section, we present the results achieved on the Bot-IoT dataset, these were omitted from the main text due to our inclusion of the Ton-IoT dataset, to which Bot-IoT is a precursor. In these experiments, we observe that among the feature selection methods leading to performance that is near optimal, FSNID selects the fewest features. Again, we note that CLM selects the fewest features. 

\section{Estimation of the Transfer Entropy Measure}\label{appendix:estimation}
Algorithm \ref{alg:estimation} illustrates the procedure used for the estimation of the transfer entropy measure.

\section{Wall Clock Runtimes}\label{app:runtimes}
In the following table, we present the wall clock runtimes for each feature set selection method in seconds.

\begin{table*}[h!]
    \centering
    \caption{Wall clock runtimes for each feature set selection method in seconds.}
    \label{tab:metrics}
    \begin{tabular}{lcccccc}
        \toprule
         & \textbf{ToN-IoT} & \textbf{NSL-KDD} & \textbf{CIC-DDoS2019} & \textbf{UNSW-NB15} & \textbf{CIC-IDS17} \\
        \midrule
        \small
        \textbf{UMFI} & 1167 & 312 & 1018 & 765 & 8883 \\
        \textbf{MIFA} & 5402 & 1933 & 56389 & 10170 & 189101 \\
        \textbf{CLM} & 3900 & 551 & 13713 & 1888 & 20001 \\
        \textbf{PI} & 198 & 167 & 304 & 389 & 2861 \\
        \textbf{LASSO} & 4.5 & 0.42 & 6.71 & 19.7 & 34 \\
        \textbf{FSNID} & 1723 & 432 & 3189 & 748 & 13565\\
        \textbf{FSNIDLSTM} &185477 &37258 & 203904& 71501 & 508093 \\
        \textbf{FSNIDTCN} & 116421 & 31928 & 171970 & 64801 & 499390\\
        \textbf{FSNIDGRU} & 121651 & 38314 & 192191 & 67647 & 507430 \\
        \bottomrule
    \end{tabular}
\end{table*}

\section{Hardware}
The classification and feature selection experiments presented in Section \ref{sec:fs_expts} were performed on a P100 Nvidia GPU cluster with 68GB of RAM. Meanwhile, the complexity analysis experiments in Section \ref{sec:comp_expts} were completed using an Apple Macbook Pro with a 2021 M1 processor, and 16GB of RAM.

\section{Table Of Top Features}\label{appendix:table}
In the tables in the following section, we report the top three features selected by each method for each dataset.
%
\begin{table*}
\centering
\caption{Top three features for Bot-IoT as selected by each method.}
\label{tab:Bot-iot}
\begin{tabular}{@{}p{2cm}p{3.5cm}p{9.5cm}@{}}
\toprule
\textbf{Dataset} & \textbf{Feature Selection Used} & \textbf{Top 3 Features} \\
\midrule
\textbf{Bot-IoT} & FSNID & Record start time; \newline Number of packets grouped by state of flows and protocols per source IP; \newline Number of inbound connections per destination IP. \\ \addlinespace
 &LSTM-FSNID& Numerical representation of feature protocol; \newline Number of packets grouped by state and protocols per source IP; \newline Dynamically detected protocols, such as DNS, HTTP, SSL. \\ \addlinespace
 & PI & Flow state flags seen in transactions; \newline Argus sequence number; \newline Numerical representation of feature protocol. \\ \addlinespace
 & CLM & Numerical representation of feature protocol; \newline Number of packets grouped by state and protocols per source IP; \newline Number of inbound connections per IP. \\ \addlinespace
 & LASSO & Numerical representation of feature protocol; \newline Average duration of aggregated records; \newline Number of inbound connections per IP. \\ \addlinespace
 & MIFA & Maximum duration of aggregated records; \newline Destination-to-source packet count; \newline Destination-to-source byte count. \\
\bottomrule
\end{tabular}
\end{table*}

\begin{table*}
\centering
\caption{Top three features for TON-IoT as selected by each method.}
\label{tab:ton-iot}
\begin{tabular}{@{}p{2cm}p{3.5cm}p{9.5cm}@{}}
\toprule
\textbf{Dataset} & \textbf{Feature Selection Used} & \textbf{Top 3 Features} \\
\midrule
\textbf{TON-IoT} & FSNID & Dynamically detected protocols, such as DNS, HTTP, SSL; \newline Source IP addresses which originate endpoints’ IP addresses; \newline Destination ports. \\ \addlinespace
 &LSTM-FSNID& Timestamp of connection for flow identifiers; \newline Destination IP addresses; \newline Values which specifies the DNS query classes. \\ \addlinespace
 & PI & Timestamp of connection for flow identifiers; \newline Recursion desired of DNS; \newline Destination ports. \\ \addlinespace
 & CLM & Source port; \newline Destination ports; \newline Missed bytes. \\ \addlinespace
 & LASSO & Duration; \newline Missed bytes; \newline Destination bytes. \\ \addlinespace
 & MIFA & HTTP Request body length; \newline HTTP Response body length; \newline DNS recursion desired. \\
\bottomrule
\end{tabular}
\end{table*}

\begin{table*}
\centering
\caption{Top three features for NSL-KDD as selected by each method.}
\label{tab:nsl-kdd}
\begin{tabular}{@{}p{2cm}p{3.5cm}p{9.5cm}@{}}
\toprule
\textbf{Dataset} & \textbf{Feature Selection Used} & \textbf{Top 3 Features} \\
\midrule
\textbf{NSL-KDD} & FSNID & Number of compromised conditions; \newline Number of bytes transferred from destination to source; \newline Service used by destination network. \\ \addlinespace
 &LSTM-FSNID& Number of failed logins; \newline Boolean value describing if a guest is logged in; \newline Number of compromised conditions. \\ \addlinespace
 & PI & Percentage of connections with same destination and IP that have activated flag REJ; \newline Number of compromised conditions; \newline Duration. \\ \addlinespace
 & CLM & Protocol type; \newline Destination host service error rate; \newline Service. \\ \addlinespace
 & LASSO & Duration; \newline Protocol type; \newline Server error rate \\ \addlinespace
 & MIFA & Same Service rate; \newline Service count; \newline Number of file creations. \\
\bottomrule
\end{tabular}
\end{table*}

\begin{table*}
\centering
\caption{Top three features for CIC-DDoS2019 as selected by each method.}
\label{tab:cic-ddos2019}
\begin{tabular}{@{}p{2cm}p{3.5cm}p{9.5cm}@{}}
\toprule
\textbf{Dataset} & \textbf{Feature Selection Used} & \textbf{Top 3 Features} \\
\midrule
\textbf{CIC-DDoS2019} & FSNID & Number of packets with at least 1 byte of TCP data carrying capacity; \newline Backwards inter-arrival time mean; \newline The max value of the inter-arrival time of the flow of packets in both directions. \\ \addlinespace
 &LSTM-FSNID& Minimum forward packet length; \newline Forward packets/s; \newline Flow packets/s. \\ \addlinespace
 & PI & Forward URG flags; \newline Forward header length; \newline Maximum forward packet length. \\ \addlinespace
 & CLM & SYN flag count; \newline Sub flow forward packets; \newline Forward packets per second. \\ \addlinespace
 & LASSO & ACK flag count; \newline Source IP; \newline Destination IP. \\ \addlinespace
 & MIFA & Packet length variance; \newline Inter-arrival time mean for a flow; \newline Forward inter-arrival time mean. \\
\bottomrule
\end{tabular}
\end{table*}

\begin{table*}
\centering
\caption{Top three features for UNSW-NB 15 as selected by each method.}
\label{tab:unsw-nb15}
\begin{tabular}{@{}p{2cm}p{3.5cm}p{9.5cm}@{}}
\toprule
\textbf{Dataset} & \textbf{Feature Selection Used} & \textbf{Top 3 Features} \\
\midrule
\textbf{UNSW-NB 15} & FSNID & Indicates the state and its dependent protocol; \newline Source bits per second;\newline Source to destination packet count. \\ \addlinespace
 &LSTM-FSNID& Forward packet length minimum; \newline Forward packet/s; \newline Flow packets/s. \\ \addlinespace
 & PI & Source TCP base sequence number;\newline Source to destination packet count;\newline Connections of the same destination address. \\ \addlinespace
 & CLM & Destination to source time to live value; \newline Source packets retransmitted or dropped; \newline Source IP address. \\ \addlinespace
 & LASSO & TCP connection setup round-trip time; \newline Destination port number; \newline Source packets retransmitted or dropped. \\ \addlinespace
 & MIFA & Represents the pipelined depth into the connection of http request/response transaction; \newline TCP connection setup time; \newline Source port number. \\
\bottomrule
\end{tabular}
\end{table*}

\begin{table*}
\centering
\caption{Top three features for CIC-IDS17 as selected by each method.}
\label{tab:cic-ids17}
\begin{tabular}{@{}p{2cm}p{3.5cm}p{9.5cm}@{}}
\toprule
\textbf{Dataset} & \textbf{Feature Selection Used} & \textbf{Top 3 Features} \\
\midrule
\textbf{CIC-IDS17} & FSNID & Destination port; \newline PSH flag count; \newline Minimum forward inter-arrival time between 2 packets. \\ \addlinespace
 &LSTM-FSNID& Destination port; \newline PSH flag count; \newline Minimum forward inter-arrival time between 2 packets. \\ \addlinespace
 & PI & Subflow Backward Bytes; \newline Total number of bytes sent forward in initial window; \newline Average forward inter-arrival time between 2 packets. \\ \addlinespace
 & CLM &  Minimum forward segment size;\newline Packet length mean; \newline Port number. \\ \addlinespace
 & LASSO & Packet Length Std; \newline Minimum forward segment size; \newline Minimum packet length. \\ \addlinespace
 & MIFA & Backward URG flags; \newline Minimum forward inter-arrival time between 2 packets; \newline Forward CWE flag count. \\
\bottomrule
\end{tabular}
\end{table*}

\end{document}